\documentclass[conference]{IEEEtran}
\usepackage{algorithm}
\usepackage{algorithmic}
\usepackage{cite}
\usepackage{amsmath,amssymb,amsfonts}

\usepackage{graphicx}
\usepackage{textcomp}
\usepackage{xcolor}
\def\BibTeX{{\rm B\kern-.05em{\sc i\kern-.025em b}\kern-.08em
    T\kern-.1667em\lower.7ex\hbox{E}\kern-.125emX}}

\usepackage{caption}
\usepackage{subcaption}

\usepackage{hyperref}
\usepackage{cleveref}

\begin{document}

\title{Performance Analysis of Convolutional Neural Network By Applying Unconstrained Binary Quadratic Programming
}

\author{\IEEEauthorblockN{1\textsuperscript{st} Aasish Kumar Sharma}
 \IEEEauthorblockA{\textit{Program in Computer Science} \\
 \textit{Georg August University of Göttingen}\\
 Goettingen, Germany \\
 email aasish.kr.sharma@gmail.com}
 \and
 \IEEEauthorblockN{2\textsuperscript{nd} Sanjeeb Prashad Pandey}
 \IEEEauthorblockA{\textit{Electronics and Computer Engineering} \\
 \textit{Institute of Engineering (IOE)}\\
 Pulchowk Campus, Lalitpur, Nepal \\
 sanjeeb77@hotmail.com}
 \and
 \IEEEauthorblockN{3\textsuperscript{rd} Julian M. Kunkel}
 \IEEEauthorblockA{\textit{Program in Computer Science} \\
 \textit{Georg August University of Göttingen}\\
 Goettingen, Germany \\
 julian.kunkel@gwdg.de}
}

\maketitle
\section*{Notice}
This work has been accepted for presentation at the IEEE COMPSAC 2025 Conference. © 2025 IEEE. Personal use of this material is permitted. The final published version will be available via IEEE Xplore at: \url{https://ieeexplore.ieee.org/}
\vspace{1em}

\begin{abstract}
Convolutional Neural Networks (CNNs) are pivotal in computer vision and Big Data analytics but demand significant computational resources when trained on large-scale datasets. Conventional training via back-propagation (BP) with losses like Mean Squared Error or Cross-Entropy often requires extensive iterations and may converge sub-optimally. Quantum computing offers a promising alternative by leveraging superposition, tunneling, and entanglement to search complex optimization landscapes more efficiently.
In this work, we propose a hybrid optimization method that combines an Unconstrained Binary Quadratic Programming (UBQP) formulation with Stochastic Gradient Descent (SGD) to accelerate CNN training. Evaluated on the MNIST dataset, our approach achieves a 10–15\% accuracy improvement over a standard BP-CNN baseline while maintaining similar execution times. These results illustrate the potential of hybrid quantum-classical techniques in High-Performance Computing (HPC) environments for Big Data and Deep Learning. Fully realizing these benefits, however, requires a careful alignment of algorithmic structures with underlying quantum mechanisms.

\end{abstract}

\begin{IEEEkeywords}
Back-Propagation (BP), Big Data Analysis, Combinatorial Optimization, Convolutional Neural Network (CNN), Deep Learning, Unconstrained Binary Quadratic Programming (UBQP).
\end{IEEEkeywords}

\section{Introduction}
\label{sec:Introduction}
The global creation, consumption, and storage of internet data have seen a monumental surge from 2010 to 2024, reaching an estimated 181 zettabytes as indicated in \cite{TotalDataVolume}. The trend underscores a significant increase of approximately 35.67\% in data usage rates over the period. The sheer volume highlights the inadequacy of human capacity for analyzing and deriving pertinent information from these data. Hence, the urge for sophisticated tools and intelligent machine development capable of effectively utilizing such vast datasets is palpable.

In the context of image processing's complexity, according to \cite{DeepLearninga}, demands a multi-stage process encompassing data cleaning, processing, analysis, manipulation, visualization, and storage in an optimal procedure for efficient subsequent processing and speedy availability. Conventional Machine Learning (ML) methods may falter with these sophisticated tasks, making the deployment of advanced intelligent machines that leverage various ML mechanisms crucial. One such specialized ML technique within the Artificial Intelligence (AI) spectrum is the Artificial Neural Network (ANN), as depicted in Figure \ref{fig:FNN}, which mimics the human brain's structure to carry out complex tasks.

Deep Learning (DL), a subset of ML, utilizes various layers of ANN or Deep Neural Networks (DNNs) to analyze and process intricate and feature-rich data. Among the different ANN models, the Convolutional Neural Network (CNN), shown in Figure \ref{fig:Typical CNN Architecture} \cite{osheaIntroductionConvolutionalNeural2015}, stands out for its image processing efficiency. The CNN model's remarkable image-analysis power lies in its ability to distinctly extract diverse image features based on input samples. The model includes layers such as convolutional, pooling, stride computation, loss and optimizer functions, fully connected, and output. These mathematical layers work in tandem, enabling the CNN to inspect different data features and make informed decisions. Notably, the CNN model employs various optimization algorithms for enhanced prediction accuracy depending on the defined objective function(s) involving maximization or minimization \cite{DeepLearninga, osheaIntroductionConvolutionalNeural2015}.

In ANN, objective functions also termed optimization functions, dictate the path—or subsequent steps—that an ANN chooses to achieve an optimal solution. The pathway depends on the maximization or minimization of the objective function(s), which might encompass multiple objectives. Back-Propagation (BP) and Simulated Annealing (SA) serve as examples of such optimization functions.

Within the DL realm, BP, represented as CNN-BP, has shown superior results albeit with slow divergence or convergence rate (high iteration counts) compared to the traditional SA method as observed by \cite{lecunBackpropagationAppliedHandwritten1989, CIT:24}. However, these methods are not without their drawbacks. Notably, they prove impractical for large datasets as pointed out by \cite{rereSimulatedAnnealingAlgorithm2015}. The SA model encounters scalability issues when working with extensively featured and large datasets. The challenges stem from SA's approaching regression and classification as a combinatorial optimization problem necessitating the exploration of various combinations and navigating local minima and maxima pitfalls, as exemplified in Figure \ref{fig:SimulatedAnnealingFlowchart}. As a result, the model faces NP-hard problem complexity when applied to such vast datasets.

This study primarily strives to address the identified limitations of the CNN-SA model, intending to create an efficient method outshining the current model, capable of surmounting its challenges and exhibiting superior performance in related applications.

The manuscript is organized into five sections: introduction (\ref{sec:Introduction}), literature review (\ref{sec:LiteratureReview}), research methodology (\ref{sec:Methodology}), analysis, results and discussions (\ref{sec:AnalysisResultAndDiscussion}), and conclusions with recommendations (\ref{sec:ConclusionAndRecommendations}). The opening section introduces the context, followed by a literature review furnishing requisite background knowledge, and a section detailing the research methodology. Analytical results and discussions present empirical findings and academic discourse, culminating with insights and recommendations. The paper concludes with relevant references, providing a coherent progression through the research topic.

\section{Literature Review}
\label{sec:LiteratureReview}
\subsection{Related Theory}
Machine Learning (ML), as described by Tom M. Mitchell \cite{mitchellMachineLearning1997}, involves a computer program improving its performance $P$ with tasks $T$ by learning from experience $E$. An array of ML techniques, such as Artificial Neural Networks (ANNs), mimic the human brain's neural structure to address tasks like signal processing and pattern recognition \cite{osheaIntroductionConvolutionalNeural2015}.

\begin{figure}
    \centering 
    \includegraphics[width=21em]{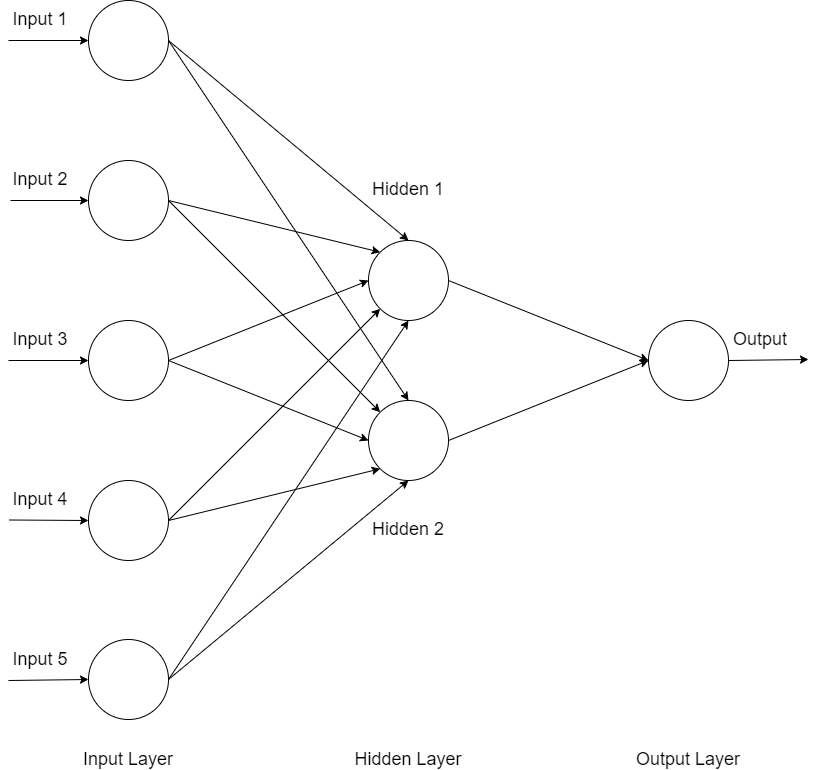} 
    \caption{The foundation of various ANN architectures, like Feed-forward Neural Networks (FNN), Restricted Boltzmann Machines (RBMs), and Recurrent Neural Networks (RNNs), is the basic three-layered feed-forward neural network (FNN) with input, hidden, and output layers \cite{DeepLearninga}.\cite{osheaIntroductionConvolutionalNeural2015}} \label{fig:FNN} 
\end{figure}

An extension of ANNs, Deep Learning (DL), includes architectures of varying depth composed of input, hidden and output layers (Figure \ref{fig:FNN} \cite{osheaIntroductionConvolutionalNeural2015}). DL involves generative models like Restricted Boltzmann Machine (RBM) and Deep Belief Network (DBN), focusing on joint probability distribution $p(x,y)$ using Bayes' Theorem. Discriminative models such as Deep Neural Network (DNN), Recurrent Neural Network (RNN), and Convolutional Neural Network (CNN) are applied mainly in supervised learning to master the conditional probability distribution $p(y
\vert x)$ \cite{osheaIntroductionConvolutionalNeural2015}.

In Big Data Analysis (BDA), ML techniques involving supervised, unsupervised, or reinforced learning process vast datasets. Model validation involves testing against actual data. Reinforcement learning, a unique form of ML, uses trial and error in a game-like setup to guide a player in achieving a goal through repeated experiences \cite{rereSimulatedAnnealingAlgorithm2015}. This intricate landscape underscores the symbiosis between DL, ML, and BDA to manage and derive insights from large datasets.

\subsubsection{Convolutional Neural Network} 
\begin{figure}
    \centering 
    \includegraphics[width=21em]{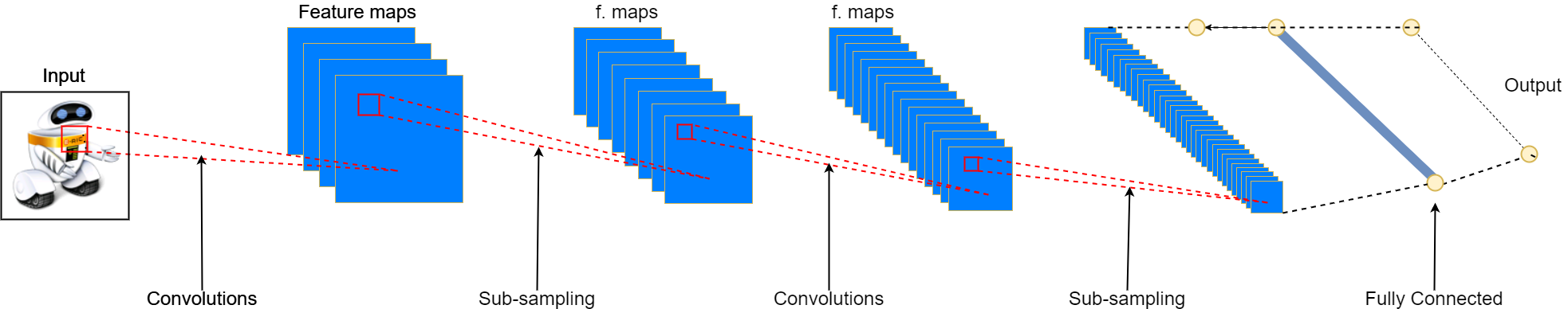} \caption{Typical CNN Architecture} 
    \label{fig:Typical CNN Architecture} 
\end{figure}

The Convolutional Neural Network (CNN), a distinguished subgroup within ANNs designed for processing medium to large feature-mapped images, is detailed in \cite{osheaIntroductionConvolutionalNeural2015}. The CNN, depicted in Figure \ref{fig:Typical CNN Architecture}, is a type of Multi-Layer Perception (MLP) model, originating from the cascading model designed by Hubel and Wiesel in the 1950s and 1960s, meant to mimic the brain's visual cells.

Developmental milestones include Yann LeCun et al.'s 1989 CNN design, marking a significant advancement within the field of computer vision \cite{lecunBackpropagationAppliedHandwritten1989}. Within this model, an image is divided, convoluted, strided, pooled, and error estimated within a single layer. Deeper configurations repeat this process across layers, depending on network depth. Predictions within the final layers use the fully connected and output layers. This study delves into two widely used objective functions for regression and classification: Mean Square Error (MSE) and Cross-Entropy (CE). Goodfellow et al.\cite{DeepLearninga} describes MSE as measuring the squared error deviation between the estimator and the true parameter, considering both bias and variance (Equation \ref{eq:mean_square_error}) \cite{osheaIntroductionConvolutionalNeural2015}.

\small \begin{equation} \centering MSE = E [(\Theta_m - \theta)^2] = Bias(\Theta_m)^2 + Var(\Theta_m) \label{eq:mean_square_error} \end{equation} \normalsize

Michael Nielsen discusses CE \cite{StudyGuideNeural, nielsenNeuralNetworksDeep2015}, which measures the deviation between the predicted probabilities and the actual output; the CE approaches zero as the neuron adeptly estimates the desired output for all training inputs (Equation \ref{eq:cross_entropy}).

\small \begin{equation} \centering CE = - \frac{1}{n} \sum_{x} {[y\space \ln \space a + (1-y)\space \ln(1-a)]} \label{eq:cross_entropy} \end{equation} \normalsize

Insightful analysis backed by visuals and mathematical formulations deepens understanding of CNNs, their mathematical underpinnings operated in image processing, and optimization.

\subsubsection{Simulated Annealing (SA)}

Simulated annealing, a stochastic meta-heuristic inspired by the metallurgical annealing process, is an optimization algorithm that maintains a probability distribution over a solution space \cite{talbiMetaheuristicsDesignImplementation2009}. This algorithm iteratively explores this space to find an optimal solution. Pioneering works have shown its applicability in wide-ranging areas such as graph partitioning and Very Large Scale Integration (VLSI) design \cite{kirkpatrickOptimizationSimulatedAnnealing1983}\cite{cernyThermodynamicalApproachTraveling1985}.

The objective function $E(x)$ represents the energy of a configuration $x$ in the solution space with the aim of this function's minimization. Techniques such as the Metropolis acceptance criterion decide the probability of accepting a new solution:

\[P (\text{accept new solution}) =min(1, e^{-\frac{\Delta E}{T}})\]

Here, $P$ denotes probability, $\Delta E$ is the energy change between the current and new solutions, and $T$ symbolizes the temperature parameter that decreases over time, controlling the likelihood of accepting worse solutions.

The Simulated Annealing (SA) algorithm (\Cref{algo:SA}), derived from the Metropolis Algorithm, follows a heuristic approach combining "divide and conquer" and "iterative improvement". This algorithm is beneficial for combinatorial searches as it reduces computation time significantly \cite{talbiMetaheuristicsDesignImplementation2009}. 

\begin{algorithm}
  \caption{SA Algorithm}
  \label{algo:SA}
  \begin{algorithmic}[1]
    \REQUIRE Cooling Schedule
    \STATE $ s = s_0 $    
    \COMMENT{Initial solution generation}
    \STATE $ T = T_{max} $    
    \COMMENT{Starting temperature}
    \REPEAT
        \REPEAT   
            \STATE Generate a random neighbor $s'$
            \STATE $\delta E = E(s') - E(s)$
            \IF{$\delta E \leq 0$}
                \STATE $ s = s' $      
                \COMMENT{Accept the neighbor solution}
            \ELSIF{$e^{-\delta E/k_BT} > \text{rand}(0,1)$}
                \STATE $ s = s' $
            \ENDIF
        \UNTIL Equilibrium conditions    
        \COMMENT{A given number of iterations executed at each temperature T}
        \STATE $ T = g(T) $   
        \COMMENT{Temperature update}
    \UNTIL Stopping criteria are met   
    \COMMENT{For example, $T<T_{min}$}
    \STATE \textbf{return} $s$
    \ENSURE The best solution
  \end{algorithmic}
\end{algorithm}

While the algorithm converges quickly, in non-ergodic systems its effectiveness decreases, and this can cause an exponential increase in estimations and the time complexity of the problem. Specifically, it may need $N$ iterations at temperature $T$ for a solution to relax after surpassing $M$ barriers, with complexity $O(n^2)$. Consideration of different loss or objective functions escalates this complexity exponentially to $O(c^n)$, resulting in NP-hard problems.

To tackle this issue, new methods are needed to lower time complexity. Inspiration for these innovative methods can come from the field of quantum physics, which represents a significant shift in methodology and aims to improve the algorithm's efficiency and scalability.

\subsubsection{Quantum Annealing (QA)}

Quantum annealing (QA) is a technique in quantum computing used to determine the global minimum of an objective function in optimization problems. While simulated annealing (SA) is also an optimization method, and both QA and SA employ adiabatic processes, QA distinctively leverages principles from quantum mechanics. In contrast to SA, where the state changes with temperature fluctuations, it is quantum fluctuations that govern state transitions in QA as seen in \Cref{algo:QA} and \ref{algo:QAt}.

\textit{Hamiltonian Process:} The intrinsic energy of a quantum system is typically described by a Hamiltonian. Within the realm of optimization, this energy pertains to the optimal solution to a problem. The equation for the Hamiltonian of a system in its ground state (or the optimized state) is as follows:

\begin{equation}
    H_v = - \frac{1}{2m}v^2 \frac{\partial^2}{\partial x^2} + V(x) 
\end{equation}

where \(v\) signifies the potential constant, \(x\) stands for the solution variable, and \(m\) is the mass of the particle (assumed as a unit mass). \(V(x)\) is the potential function encoding the cost function for minimization \cite{mukherjeeMultivariableOptimizationQuantum2015a}. 

As per a linear interpolation between an initial Hamiltonian \(H_0\) and a final Hamiltonian \(H_p\), the time-dependent Hamiltonian \(H(t)\) is defined as:

\begin{equation}    
    H(t)=A(t)H_0+B(t)H_p
\end{equation}

where \(A(t)\) and \(B(t)\) are annealing schedules dictating the transition from the initial to the final Hamiltonian over a time period. The Schrödinger equation drives the time evolution of the quantum state:

\begin{equation}
    i\hbar\frac{d}{dt} |\Psi(t)\rangle = H(t) |\Psi(t)\rangle 
\end{equation}

In an optimized state, the quantum system reaches the ground state of the final Hamiltonian. Like SA, QA uses quantum fluctuations analogous to the temperature in SA. However, while SA employs thermal jumps to avoid local minima, QA uses quantum tunneling. The quantum nature of QA modifies non-ergodic systems into ergodic forms, making it a more efficient optimization technique than SA \cite{CIT:19}. 

The transition of quantum states is represented as:

\begin{equation}
    H(t) = H_0 + \lambda(t)H',
\end{equation}

With \(H_0\) as the classical Hamiltonian and \(H'\) as the quantum transition between states. When the time-dependent quantum kinetic term \((\lambda(t)H')\) is added, the total Hamiltonian of the system becomes \(H(t)\). The evolution of the system is governed by solving the time-dependent Schrödinger equation:

\begin{equation}
     i\hbar \frac{\partial\Psi }{\partial t} = [\lambda(t)H' + H_0]\Psi ,
\end{equation}

where \(|\Psi(t)|\) represents the state of the system at time \(t\) and \(\hbar\) symbolizes the reduced Planck constant. 

\begin{algorithm}
  \caption{QA Algorithm -- Procedure 1: Quantum Annealing}
  \label{algo:QA}
  \begin{algorithmic}[1]
    \REQUIRE Initial condition \(init\), control parameter \(v\), duration \(t_{max}\), 
              tunnel time \(t_{drill}\), local optimization time  \(t_{loc}\).
    \STATE \(t \leftarrow 0\)
    \STATE \(\epsilon \leftarrow init\)
    \STATE \(v_{min} = \text{cost}(\epsilon)\)
    \WHILE{\(t < t_{max}\)}
        \STATE \(j \leftarrow 0\)
        \REPEAT
            \STATE \(i \leftarrow 0\)
            \REPEAT
                \STATE \(\epsilon \leftarrow \text{Quantum Transition}(\epsilon, v, t_{max})\)
                \IF{\(\text{cost}(\epsilon) < v_{min}\)}
                    \STATE \(v_{min} \leftarrow \text{cost}(\epsilon)\)
                    \STATE \(i, j \leftarrow 0\)
                \ELSE 
                    \STATE \(i \leftarrow i + 1\)
                \ENDIF
            \UNTIL \(i > t_{loc}\)
            \STATE \(\epsilon \leftarrow \text{Local Optimization}(\epsilon)\)
            \IF{\(\text{cost}(\epsilon) < v_{min}\)}
                \STATE \(v_{min} \leftarrow \text{cost}(\epsilon)\)
                \STATE \(j \leftarrow 0\)
            \ENDIF
        \UNTIL \(j < t_{drill}\)
        \STATE Draw a trajectory of length \(Vt_{max}\) and jump there.
        \STATE \text{Local Optimization} \((\epsilon)\)
    \ENDWHILE
    \ENSURE \text{Local Optimization.}
  \end{algorithmic}
\end{algorithm}

\begin{algorithm}
  \caption{QA Algorithm -- Procedure 2: Quantum Transitions}
  \label{algo:QAt}
  \begin{algorithmic}[1]
    \REQUIRE Initial condition \( \epsilon \), chain length \( vt \), set of neighbors to estimate \( N eigh \).

    \FORALL{neighbor \( k \in N eigh \)}
        \STATE Estimate the wave function \( \Psi_v(k) \);
    \ENDFOR
    \STATE \( best \leftarrow \) select a neighbor in \( N eigh \) with the probability proportional to \( \Psi_v \)
    \RETURN \( best \)
    \ENSURE The best solution is found.
  \end{algorithmic}
\end{algorithm}

\subsubsection{Simulated Annealing (SA) Versus Quantum Annealing (QA)}
\begin{figure}
\centering
    \includegraphics[width=21em]{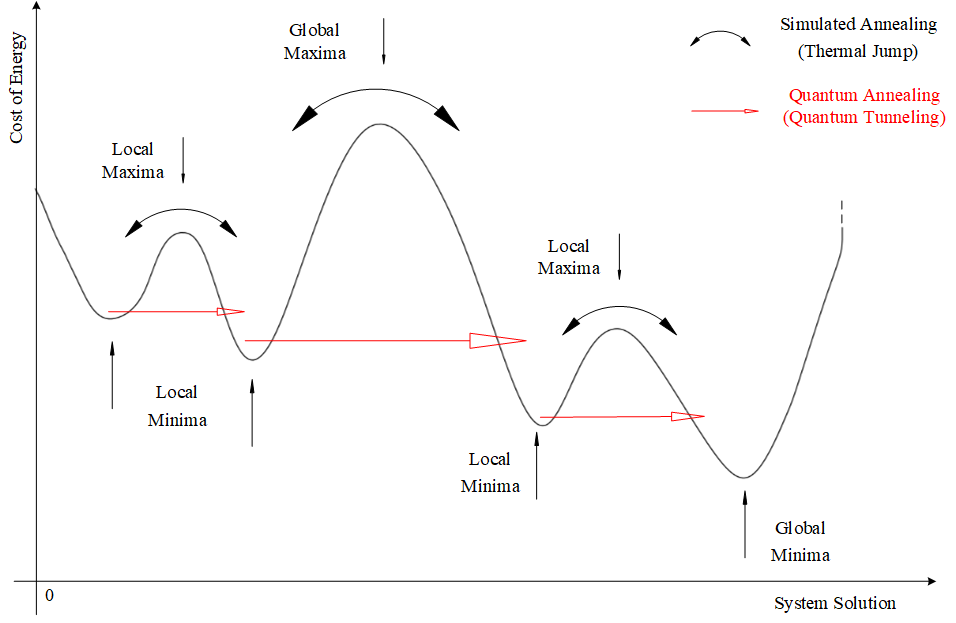}
    \caption{Simulated Vs Quantum Annealing}
    \label{fig:SimulatedVsQuantumAnnealing}
\end{figure}
The differences between QA and SA are illustrated in Figure \ref{fig:SimulatedVsQuantumAnnealing}. 
\begin{itemize}
    \item SA, a classical method derived from annealing in physics, operates within the constraints of classical physics using controlled temperature, whereas QA operates within the constraints of quantum physics using quantum fluctuations \cite{CIT:3_article_Simulated_Annealing_Algorithm_for_Deep_Learning,CIT:6,CIT:22_article_Simulated_Quantum_Annealing_Can_Be_Exponentially_Faster_than_Classical_Simulated_Annealing}.
    \item QA leverages quantum superposition and entanglement to explore multiple possibilities at the same time, whereas SA probabilistically explores classical states.
    \item QA excels in identifying local and global energy minima in optimization problems, and efficiently samples lower energy states in sampling problems, making it valuable in machine learning applications.
\end{itemize}

\par
\subsubsection{Quantum Approach for CNN}

There are two primary approaches to adopting a CNN model via quantum annealing: (i) the Quantum Universal Gate (QUG) approach, and (ii) the Classical-Quantum Hybrid (CQH) approach \cite{CIT:0_book_Deep_Learning,CIT:6,CIT:21,CIT:22_article_Simulated_Quantum_Annealing_Can_Be_Exponentially_Faster_than_Classical_Simulated_Annealing,CIT:25,CIT:26,CIT:31}. These approaches integrate classical convolutional neural networks with quantum annealing, opening avenues of exploration for improved computational strategies and potential enhancements in optimization tasks.

\vspace{0.35em}

(i) \textit{Quantum Universal Gate (QUG) Approaches:} In this approach, a desired neural network is formulated using quantum circuits and quantum gates. The methods used in this approach include Quantum Circuit Learning, Parameterized Quantum Circuits, Quantum Neural Networks, Quantum Gradient Descent, and Variational Quantum Circuits. The Quantum Convolutional Neural Network (QCNN) utilizes the quantum circuit model and employs the Multi-scale Entanglement Renormalization Ansatz (MERA) tensor network for efficiently describing critical systems \cite{CIT:31}. 

\vspace{0.35em}

(ii) \textit{Classical-Quantum Hybrid (CQH) Approaches:}\label{sssec:II_CQH_Approach}  These methods aim to design a quantum neural network through the application of adiabatic techniques like Quantum Annealing. The approaches used under this category include Quantum Annealing (QA), Quantum Approximate Optimization Algorithm (QAOA), Quantum Support Vector Machine (QSVM), Quantum Kernel Methods (QKM), and Quantum Principal Component Analysis (QPCA). Additional hybrid techniques based on quantum data encoding are Amplitude Encoding (AE), Basis Encoding (BE), Parameter Encoding (PE), and Quantum Feature Maps (QFM).

Studies integrating RBM architecture with a traditional variational Monte Carlo method using paired-product (geminal) wave functions have indicated enhanced accuracy in computing ground-state energy \cite{jiaQuantumNeuralNetwork2019}. Additionally, a proposed hybrid CNN model consists of an input layer, a convolutional layer, a pooling layer, and a quantum annealing layer. This model uses the Software Development Kit (SDK) provided by D-Wave and is compatible with a classical computing environment \cite{CIT:19, CIT:22_article_Simulated_Quantum_Annealing_Can_Be_Exponentially_Faster_than_Classical_Simulated_Annealing}. 

\subsubsection{Quadratic Unconstrained Binary Optimization (QUBO) Model}

The QUBO model is useful for a wide range of combinatorial optimization problems. It allows the transformation of disparate objective functions into matrix form suitable for Quantum Annealing and Quantum Approximate Optimization Algorithm. The choice of annealing process primarily depends on the Qubit size and network design, typically using Chimera or Pegasus connection topologies. The DW2000Q5/6 QPU system, known for its low noise levels, can scale up to 2030 Qubits \cite{CIT:25, CIT:26}.

\subsection{Related Work}

\begin{figure}
    \centering
    \includegraphics[width=21em]{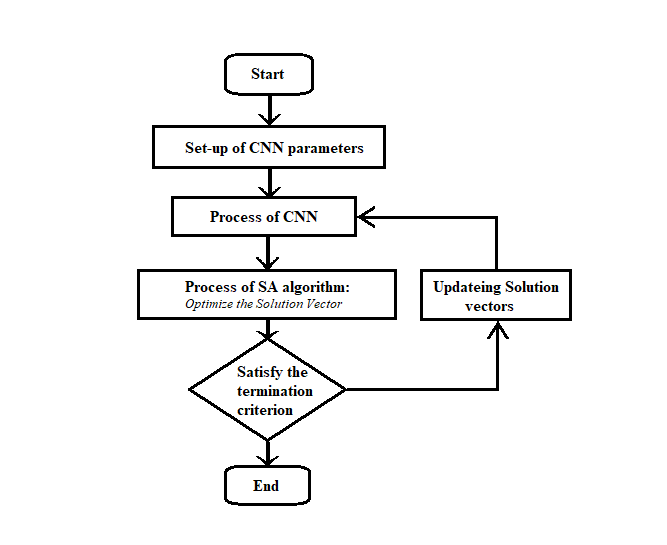}
    \caption{Optimization Flowchart for CNN Using Simulated Annealing Algorithm: A Proposal by L.M. Rasdi Rere et. al. (2015) \cite{rereSimulatedAnnealingAlgorithm2015}.}
    \label{fig:SimulatedAnnealingFlowchart}
\end{figure}

- \textit{CNN Optimization Using SA:}
The paper investigates the improvement of CNN performance by harnessing SA in deep learning with the MNIST dataset. It acknowledges the trade-off between performance improvement and computation time when comparing the enhanced method to traditional CNN with back-propagation. In an associated paper, "Simulated Annealing Algorithm for Deep Learning", L.M. Rasdi Rere et al. \cite{CIT:3_article_Simulated_Annealing_Algorithm_for_Deep_Learning} utilized SA in fine-tuning a CNN model using MSE as one of the fitness functions. This revolutionary work informs our current study by offering noteworthy insights about applying SA in deep learning. The research method is depicted in \Cref{fig:SimulatedAnnealingFlowchart}.

- \textit{Power of Quantum Computing: Future of Scientific Progress:}
The potential of quantum computing in meeting the expanding computational needs of various departments, especially in U.S. Energy and Scientific sectors, was emphasized in the "ASCR Report on Quantum Computing for Science" \cite{CIT:5}. The report indicated that machine learning stands to gain remarkably from quantum computing technologies, marking a crucial scientific breakthrough.

- \textit{Insights from QA Application in Deep Neural Network Training:}
A pathbreaking work by Adachil and Henderson called "Application of Quantum Annealing to Training of Deep Neural Networks" \cite{CIT:4} demonstrated superior accuracy in training Deep Belief Network Models (DBNM) using Quantum Annealing Machine (QAM) over the classical Restricted Boltzmann Machine (RBM). Their research supplies vital insights into the application of quantum computing in the training of deep neural networks.

- \textit{State of CNN: Deep Learning \& Quantum computing:}
Research by \cite{CIT:2} concentrates on supercharging CNN with robust GPUs and Quantum Computing methods. They proposed creative solutions, such as QA for efficient weight determination, high-performance computing for identifying optimal network configurations, and neuromorphic computing for better cost and power performance on complex topologies. Their study affirms previous observations that Quantum Computers can considerably streamline CNNs \cite{CIT:2, CIT:3_article_Simulated_Annealing_Algorithm_for_Deep_Learning, CIT:4}.

- \textit{Challenges \& Opportunities in QA and Quantum Neural Networks (QNN):}
Despite UQGM's versatility, modeling with universal quantum gates confronts challenges connected to quantum state transitions and bespoke configurations \cite{CIT:1, CIT:8, CIT:9, CIT:19, CIT:20, CIT:25}. Yet, QA shows promise in dealing with discrete search space characteristics, making it useful with generative models like DBM and DNN \cite{CIT:36}. In 2019, Iris et al. proposed Quantum Convolutional Neural Networks (QCNN) detailing efficient training that could potentially aid in near-term quantum device implementation \cite{CIT:31}.

- \textit{Bridging the Digital-Adiabatic Gap - Integrating QA into Classical CNN Models:}
The foremost question considered is how to use QA efficiently in enriching classical CNN models. To grapple with this, a pioneering method proposes the introduction of a QA layer immediately after the loss function estimation, aiding in merging digital and adiabatic processes seamlessly (\Cref{fig:Flowchart of QA Method in Details}).

- \textit{Garnering Insights from Expert Communities:} 
\label{sssec:ExpertCommunities}
Interactions on platforms such as ResearchGate \cite{CIT:29} and StackExchange \cite{CIT:30} furnished valuable insights into our study. The shared expertise highlighted the critical step of formulating the optimization function into a QUBO matrix for effectively implementing QA in the training of CNN models, particularly with D-Wave's annealer.

\par

- \textit{Challenges and Considerations:}    
Aligning quantum annealing and CNN models promises exciting synergies, yet uncertainties remain, especially regarding the Universal Quantum Gate Model application. Experts caution that formulating solutions for combinatorial optimization problems using a universal gate model may demand significant computational resources. The optimality of such an approach for large combinatorial states is a concern, underscoring the importance of careful assessment of trade-offs associated with various quantum computing models.

\subsection{Why a Hybrid QA Approach?}
\label{ssec:Reasons_for_Applying_Hybrid_QA_approach}
The choice of a hybrid QA approach over other available QUG and CQH approaches (\Cref{sssec:I_QUG_Approach}) is guided by the following reasons:

\subsubsection{Supported by the Professional Community}
A related inquiry on StackOverflow (\Cref{sssec:ExpertCommunities}) elicited suggestions from community experts who recommended QA as a suitable optimization option for our outlined SA problems.

\subsubsection{Qubits Requirement} 
In QUG-based Approaches (\Cref{sssec:I_QUG_Approach}), the qubit count required corresponds with the total parameters (weights and biases) in the neural network. For example, A simple encoding for MNIST images on a quantum computer could require approximately 6272 qubits, assuming each pixel's grayscale value is represented by an 8-bit binary string. This rough estimation shows the model would need a significantly high number of qubits exceeding the count available for open-sourced systems. In contrast, the hybrid CQH approach \Cref{sssec:II_CQH_Approach}, such as QA, requires fewer qubits as it only applies to loss functions, with the classical part handling neural network parameter optimization. This significant reduction in required qubits makes the hybrid approach more feasible.

\subsubsection{QA Vs. QAOA} 
Recent comparative studies (\cite{pelofskeQuantumAnnealingVs2023}, \cite{pereraLexicalizingLinkedData2017a}) between QA and QAOA (CQH approaches, \Cref{sssec:II_CQH_Approach}) suggest that QA outperforms QAOA. The higher error correction level and the additional layers in QAOA that slow down the process might be the reason why classical computers find it more challenging to simulate the QUG approach compared to the CQH (QA) approach.
\par

\section{Research Methodology}
\label{sec:Methodology}
\subsection{Proposed Research Methodology: Chronological Overview}
To accomplish our stated research objectives, we adopt a systematic and chronological methodology, presented in a structured approach that allows for an organized investigation. The subsequent sections describe the integral phases and processes involved in this research methodology.

\begin{figure}
\centering
\includegraphics[width=0.5\textwidth]{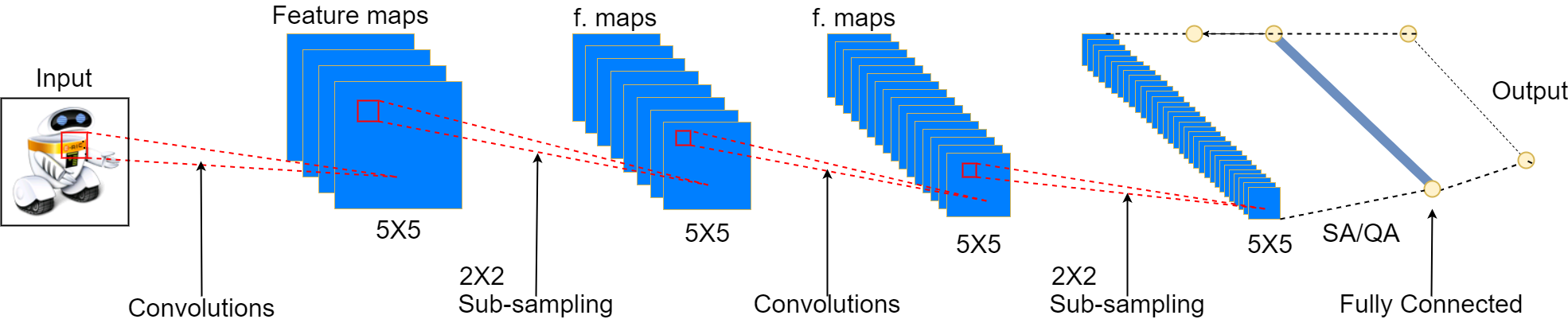}
\caption{Diagrammatic representation of proposed methodology}
\label{fig:Diagrammatic_representation_of_proposed_research_methodology}
\end{figure}

Figure \ref{fig:Diagrammatic_representation_of_proposed_research_methodology} shows how the input image of $28 \times 28$ undergoes convolution with a $5 \times 5$ kernel, followed by a $2 \times 2$ pooling. This process is repeated twice, corresponding to the layer count in the model.

In the fully-connected layer, the model uses an optimization function - either SA or QA - to facilitate the convergence of the objective function, similar to the path traced by the Stochastic Gradient Descent (SGD). This also aids in propagating the loss through the network weights back to the initial node, allowing continual optimization of the CNN model. Here is the step-by-step procedure.

\subsection{Procedure for Analyzing CNN with SA}
\par
(i) \textit{Model Preparation:} The configuration and testing of the CNN model with SA optimization draw from methods outlined by Rere L.M. et al. \cite{CIT:3_article_Simulated_Annealing_Algorithm_for_Deep_Learning}.
\par
(ii) \textit{Model Structure:} The model comprises an input layer, convolution layers with corresponding pooling layers, and a final output or fully-connected layer, maintaining the structure pursued in the referenced paper. The chosen optimization loss functions are the MSE and CE.
\par
(iii) \textit{Optimization Process:} We start by calculating CNN weights and biases. Then, the SA method facilitates annealing the loss function to obtain the optimal solution, while updates to the solution vector follow either through a random value addition $\Delta x$ or by adhering to the Boltzmann distribution, as defined in SA Algorithm \ref{algo:SA}.
\par
\begin{figure}
\centering
\includegraphics[width=0.5\textwidth]{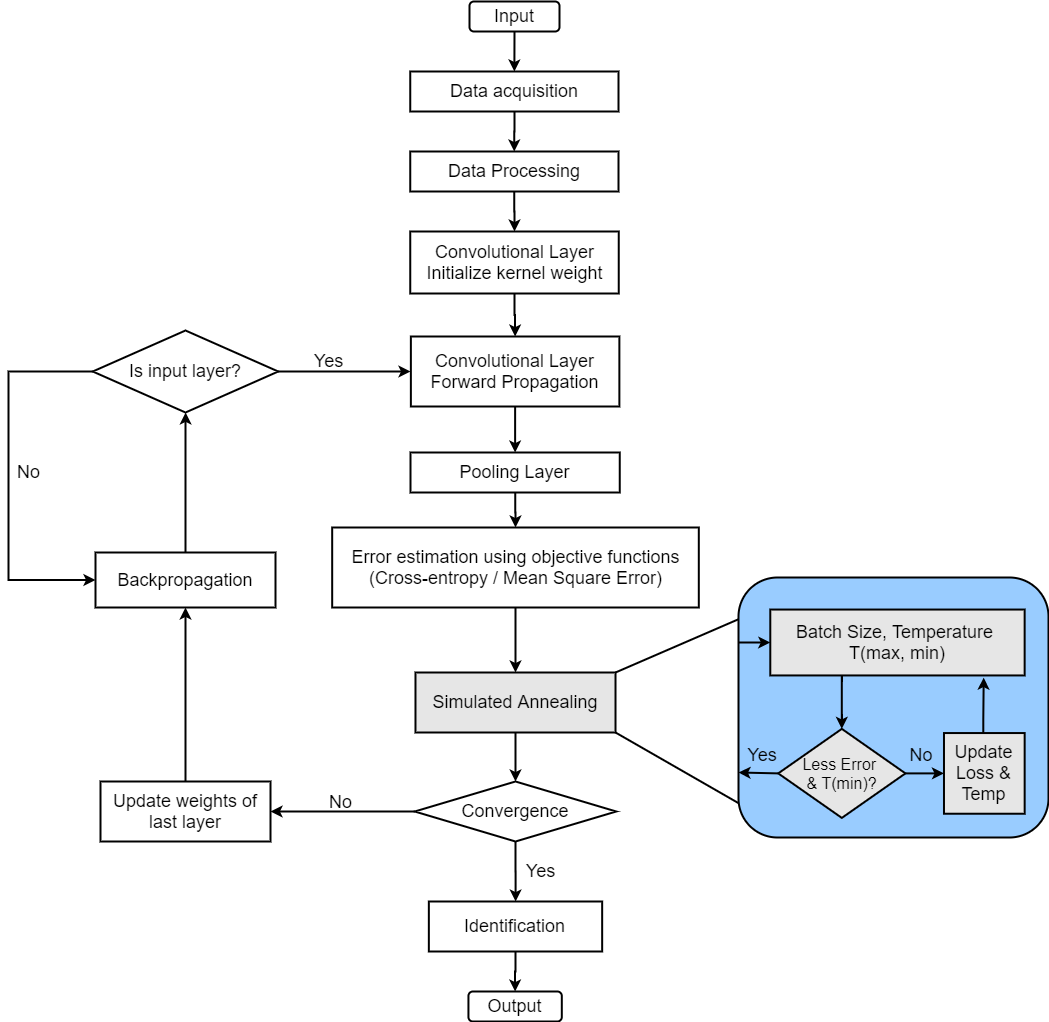}
\caption{Flowchart of the SA Method}
\label{fig:Flowchart of SA Method in Details}
\end{figure}

(iv) \textit{Updating Weights and Biases:} Upon obtaining the optimal solution, all layers' weights and biases are updated using the SGD method, as demonstrated in Figure \ref{fig:Flowchart of SA Method in Details}.
\par
(v) \textit{Performance Evaluation:} Although the CNN-SA model provides improved optimization advances over back-propagation, it presents considerable time and resource challenges. Our study targets optimizing the CNN-SA system that minimizes time costs during training while maintaining classification accuracy.

\par

\subsection{Procedure for Analyzing CNN with QA}

\par
(i) \textit{Preparation of the QA Model:}
This step involves the configuration and testing of the CNN model using QA within DWAVE's framework.
\par    
(ii) \textit{Optimization Process:}
Significant alterations are introduced in this stage, substituting QA for SA. This change necessitates the conversion of the loss function into a QUBO matrix through the natural QUBO formulation method for minimization matrices \cite{CIT:25}.
\par
\begin{figure}
\centering
\includegraphics[width=0.5\textwidth]{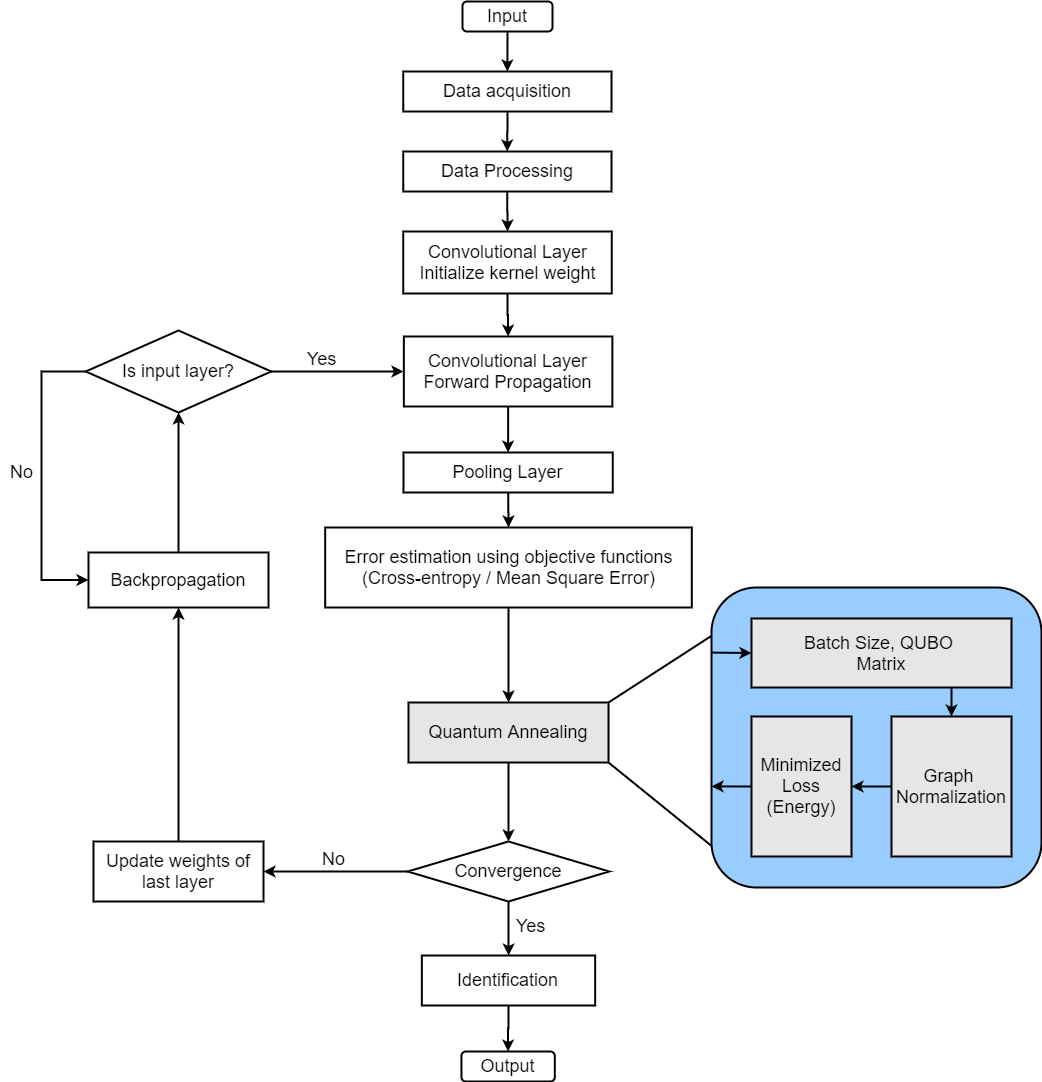}
\caption{Flowchart of the QA Method}
\label{fig:Flowchart of QA Method in Details}
\end{figure}

    
(iii) \textit{Execution of Quantum Annealer:}
The constructed QUBO matrix is then supplied to DWAVE's quantum annealer. By following quantum behaviours, the QA process provides an optimal estimated solution as depicted in the flowchart on \Cref{fig:Flowchart of QA Method in Details}.
\par
(iv) \textit{Further Elaboration:}
Additional details regarding the QA process, QUBO formulation considerations, and the specifics of DWAVE's quantum annealer, are expounded in subsequent sections to provide a comprehensive understanding of Quantum techniques, such as Annealing applied to the CNN model.

\subsubsection{Procedure for Mapping Parameters to Qubits}
The accurate mapping of parameters to qubits is vital in problem optimization encoding for quantum computers or quantum annealers. This study employs QA in a novel fashion to enhance the efficiency of CNN training. The first step is to convert the loss function of CNN into a QUBO form following the methodology by Glover et al. \cite{CIT:25}. This QUBO formulation integrates smoothly with DWAVE's annealing model, initiating QA to minimize the cost function. This minimized cost value is crucial in estimating the minimal cost/error during each CNN training batch iteration. Following QA, Stochastic Gradient Descent (SGD) optimizes the weights and biases of the network in line with the CNN training steps, demonstrating an innovative blend of quantum and classical optimization techniques.

- \textit{QUBO Formulation:}
The formulation of the loss function into QUBO follows the tutorial by Glover et al. \cite{CIT:27}. The function is expressed as a complete symmetric Q matrix:
\small\begin{equation}
    QUBO: minimize/maximize y = x^tQx
    \label{equ:QUBO}
\end{equation}\normalsize
where $ x $ is a vector of binary decision variables and $ Q $ is a square matrix of constants. The $ Q $ matrix can be in \textit{symmetric} or \textit{upper triangular} form.
For the $ Symmetric form $: For all $i$ and $j$ except $i$ = $j$, replace $ q_{ij} $ by $  (q_{ij} + q_{ji})/2 $,
For the $ Upper triangular form $: For all $ i $ and $ j $ with $ j > i $, replace $ q_{ij} $ by $ q_{ij} + q_{ji} $, then replace all $ q_{ij} $ for $ j<i $ by 0. If the matrix is already symmetric, this doubles the $ q_{ij} $ values above the main diagonal and then sets all values below the main diagonal to 0 \cite{CIT:25}.

\par

\subsubsection{Qubit Encoding Procedure for MSE and CE Hamiltonian Objective Functions via QUBO}

The following steps illustrate the process of encoding Hamiltonian objective functions of MSE and CE losses into qubits using the QUBO formulation:

\begin{enumerate}
    \item \textit{Define MSE and CE Errors:} The formula representing MSE and CE losses for a machine learning model parameter set \(\theta\) are as follows:
    \begin{equation}
         MSE(\theta)=\frac{1}{N}\Sigma_{i=1}^N{(yi-f(xi,\theta))}^2
    \end{equation}
    \small
    \begin{equation}
            CE(\theta) = - \frac{1}{N} \Sigma_{i=1}^N (y_i \log (f(x_i, \theta)) + (1 - y_i) \log (1 - f(x_i, \theta)))
    \end{equation}
    \normalsize
    \item \textit{Define Binary Variables $x_i$:} Each parameter and its potential values is represented by this binary variable.
    
    \item \textit{Transform into an Ising Hamiltonian:} At this stage, the Ising Hamiltonian is formulated as: 
    \begin{equation}
        H(\sigma)=\Sigma_{i=1}^N a_i\sigma_i + \Sigma_{i<j}^N b_{ij}\sigma_i \sigma_j    
    \end{equation}
    
    \item \textit{Define Bias Terms ($a_i$):} MSE bias terms are defined as: 
    \small$a_i=\frac{2}{N}(f(x_i,\theta) - y_i)$\normalsize, and CE bias terms use: 
    \small$a_i=-\frac{1}{N}(y_i-f(x_i,\theta))$\normalsize.
    
    \item \textit{Establish Coupling Terms ($b_{ij}$):} For MSE, coupling terms are defined as:
    \small$b_{ij}=\frac{4}{N^2}(f(x_i,\theta) - y_i)(f(x_j,\theta) - y_j)$\normalsize, and for CE, coupling terms use:
    \small$b_{ij}=-\frac{1}{N^2}(y_i-f(x_i,\theta))(y_j-f(x_j,\theta))$\normalsize.
    
    \item \textit{Assign Parameters to Qubits:} At this stage, each parameter $\theta$ is assign to a particular qubit.
    
    \item \textit{Encode Qubit Interaction Coefficients:} The coefficients $a_i$ and $b_{ij}$ are encoded for qubit interactions.
    
    \item \textit{Formulate QUBO Objective Function:} The QUBO objective function is formed using the following equation:
    \small$\text{QUBO}=\Sigma_i q_ix_i+ \Sigma_{i<j} q_{ij}x_ix_j$\normalsize.
    
    \item \textit{Map Parameters to Binary Variables:} Each binary variable; 0 or 1, represents a specific parameter value.
    
    \item \textit{Set QUBO Coefficients:} The coefficients $q_i$ and $q_{ij}$ in the QUBO formulation are set based on pre-calculated values.
    
    \item \textit{Run Quantum Annealing Optimization Algorithm:} The QUBO is executed on a quantum annealer to find the values that minimize the objective function.
    
    \item \textit{Decode Binary Variables:} The binary variable values are decoded after the quantum annealer optimization to obtain minimized parameters.
    
    \item \textit{Post-Processing:} Results are analyzed, decoded values are interpreted, and the solution is refined if needed.
\end{enumerate}

\textit{Note:} The effectiveness of this method depends on the optimization problem and the capabilities of the quantum annealer, as well as the hardware details of the specific quantum computing or classical optimization tool.

\subsubsection{Designing a Hybrid CNN-QA Model: Integration of Formulated QUBO within the CNN Layers}

The formulated QUBO matrix is integrated within the convolutional and pooling layers of a CNN model. Here, DWAVE's quantum annealer is employed to minimize energy to achieve the target minimization function. This procedure is repeated iteratively across all data batches. Upon training completion, the final output from the model is recorded.

To ensure the real hardware's compatibility, the code was tested using DWAVE's 'Leap' QA machine during an available free time-slot. After a successful test, the one-minute free Quantum Processing Unit (QPU) execution time was fully utilized in small-scale tests. Due to this time constraint for real-time QA, subsequent more extensive evaluations were conducted on an in-house system using DWAVE's 'neal' simulator.

%

\section{Analysis, Results, and Discussion}
\label{sec:AnalysisResultAndDiscussion}

Our research methodology and data collection resulted in findings obtained through rigorous mathematical and statistical analyses. Utilizing validation and verification processes, we were able to streamline the data into concise outcomes.

\subsection{Validation and Verification}

Following the testing and evaluation program by National Institute of Standards and Technology (NIST)\footnote{\url{https://www.nist.gov/}}, we performed validation and verification procedures for two algorithms by comparing them with a reference model. These procedures involved conducting cross-validation tests and 10-fold cross-validation tests on a 60,000-image MNIST dataset, measured against BP applied CNN-SA and CNN-QA reference models \cite{CIT:32,CIT:33}. We opted for the 10-fold cross-validation since it is widely accepted in machine learning for its efficiency in estimating model performance. Given the comparatively smaller dataset and our aim to observe advancements with fewer iterations, we decided to limit the process to 10 epochs. The evaluation centered on the classical baseline BP and CNN-SA models, with key points of interest being validation and test accuracy, execution time for 10-fold cross-validation, a 10-epoch test, and additional tests for various dataset sizes.





\subsection{Tests Conducted}

For a comprehensive analysis, we carried out a comparative study between the standard CNN model using Back-Propagation (CNN-BP) and the CNN models leveraging SA (CNN-SA) and QA (CNN-QA) methodologies. The conducted tests entailed:

\begin{itemize}
    \item \textit{10-Fold Cross-Validation Test:} We performed a comparative assessment of all three CNN models through a 10-fold cross-validation approach employing Cross-Entropy (CE) and Mean Squared Error (MSE) objective functions. Each test was executed for a single epoch with learning rates of 1, 0.1 and 0.01.
    \item \textit{10 Epoch Test:} The CNN models were evaluated across ten epochs, using a learning rate of 0.1 for both CE and MSE objective functions. Since BP tends to yield better accuracy with lower learning rates, we tested it with a slightly higher rate.
    \item \textit{Different Data Sizes Test:} The full dataset of 60,000 records were segmented into cumulative subsets, ranging from 10,000 to 50,000, to study how the performance evolves with increasing data size. The tests were performed for a single epoch, using a learning rate of 0.1, and encompassing both CE and MSE objective functions.
\end{itemize}

The purpose of these tests were to gather insights about the performance of models across different hyprer-parameters, datasets and sizes \cite{CIT:32,CIT:33}.

\subsection{System Specifications}


\textit{Processor:} Intel(R) Core(TM) i5-3230M, 2.60 GHz, \textit{Memory (RAM):} 8 GB, \textit{System Drive (Storage):} SSD, \textit{Operating System:} Windows 10, 64-bit, \textit{Programming Language:} Python 3.7+, 
\textit{Development Tool:} PyScripter, \textit{Python Packages:} Numpy, sklearn, matplotlib, D-Wave SDK-Neal, D-Wave-QbSolv, and \textit{Database:} MNIST, 60,000 open-sourced standard images of size $28 \times 28$ for the train set and 10,000 for the test set.

\subsection{Observations}    
\begin{figure}
    \centering
    \begin{subfigure}[b]{0.5\textwidth}
        \includegraphics[width=0.45\textwidth]{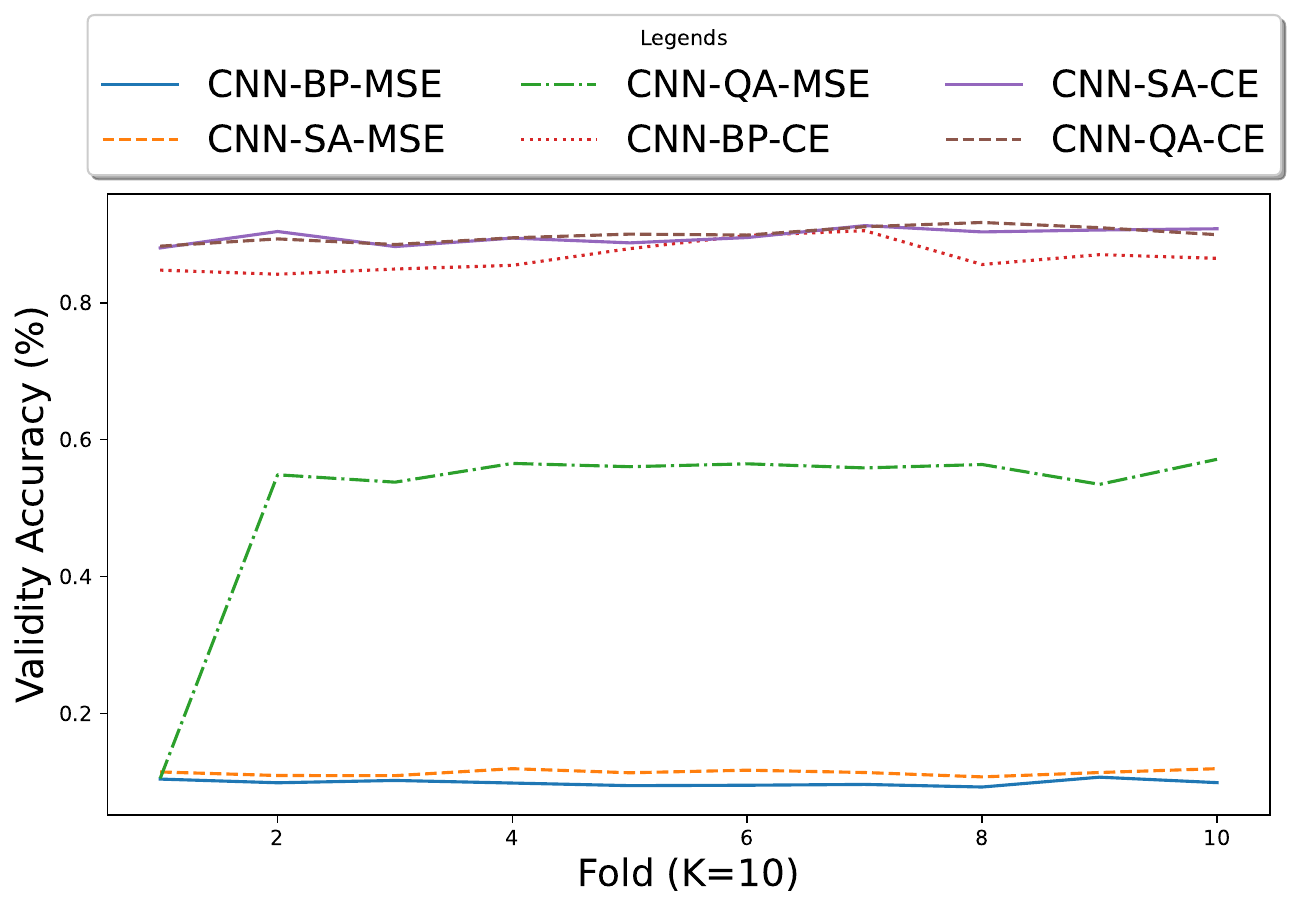}      
        \includegraphics[width=0.45\textwidth]{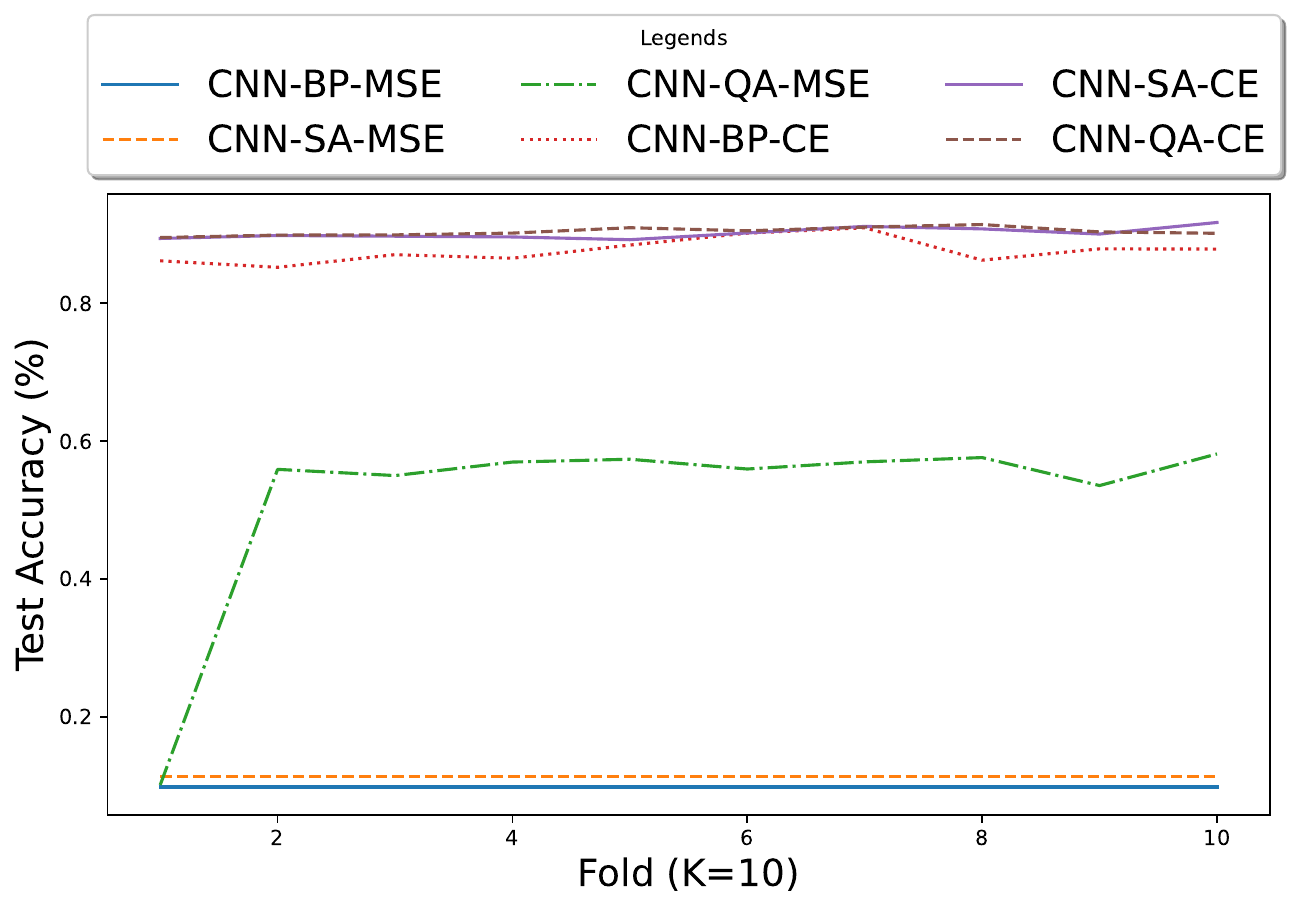} 
        \caption{Validation and Test Accuracy (\%) - Line Graph}
    \end{subfigure}
    
    \begin{subfigure}[b]{0.5\textwidth}
        \includegraphics[width=0.45\textwidth]{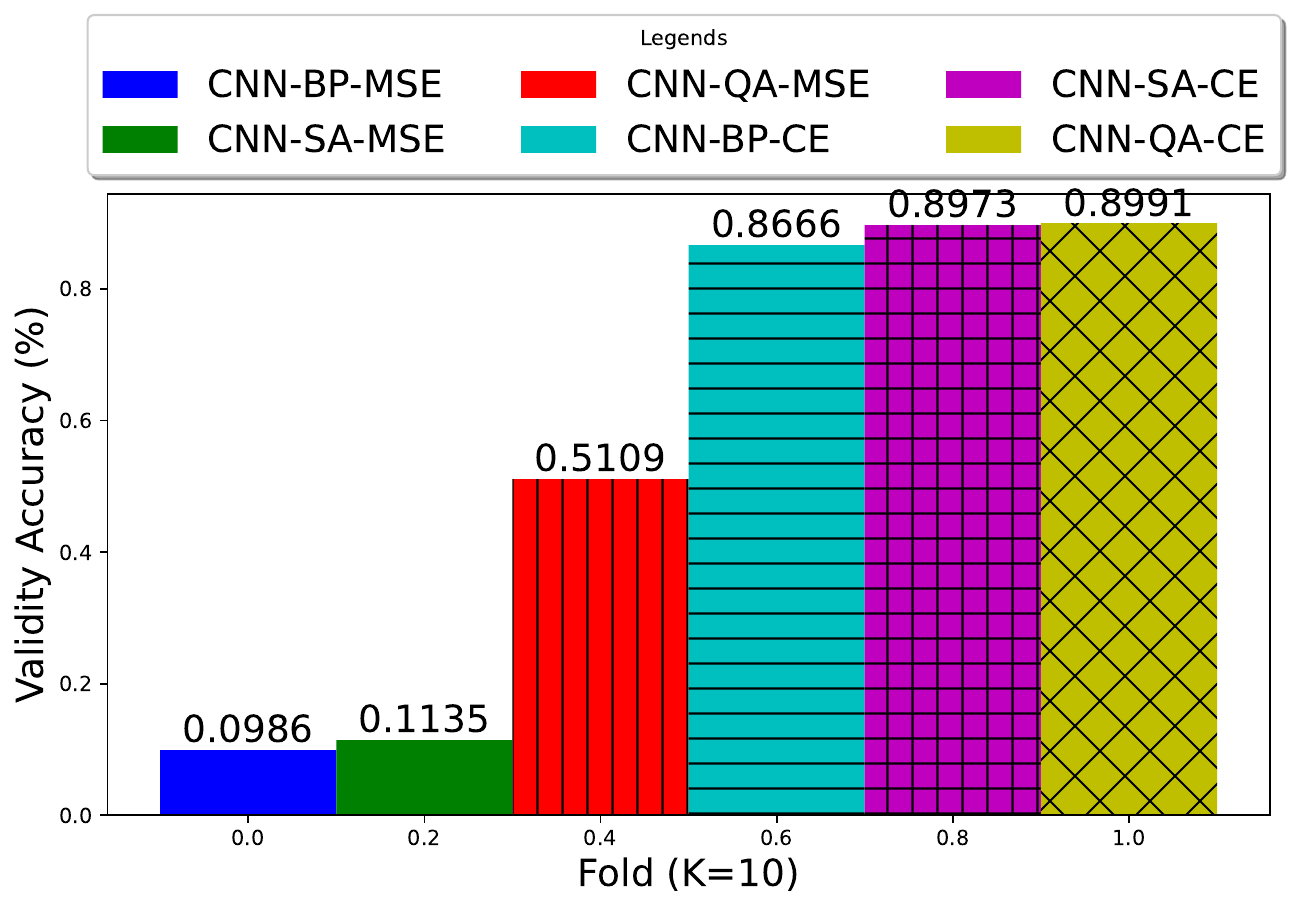}  
        \includegraphics[width=0.45\textwidth]{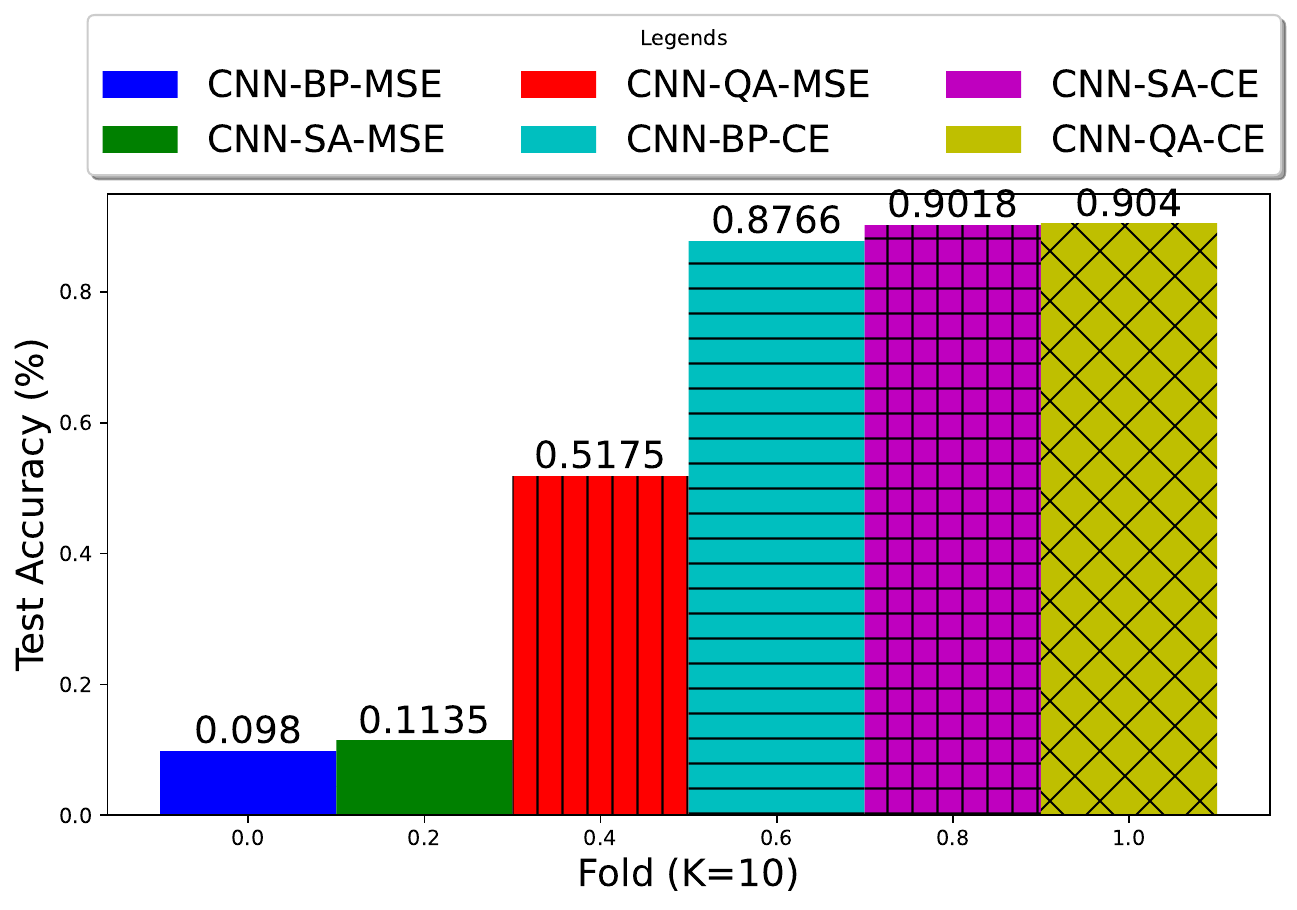} 
        \caption{Validation and Test Accuracy (\%) - Bar Graph}
    \end{subfigure}
    
    \begin{subfigure}[b]{0.5\textwidth}
        \includegraphics[width=0.45\textwidth]{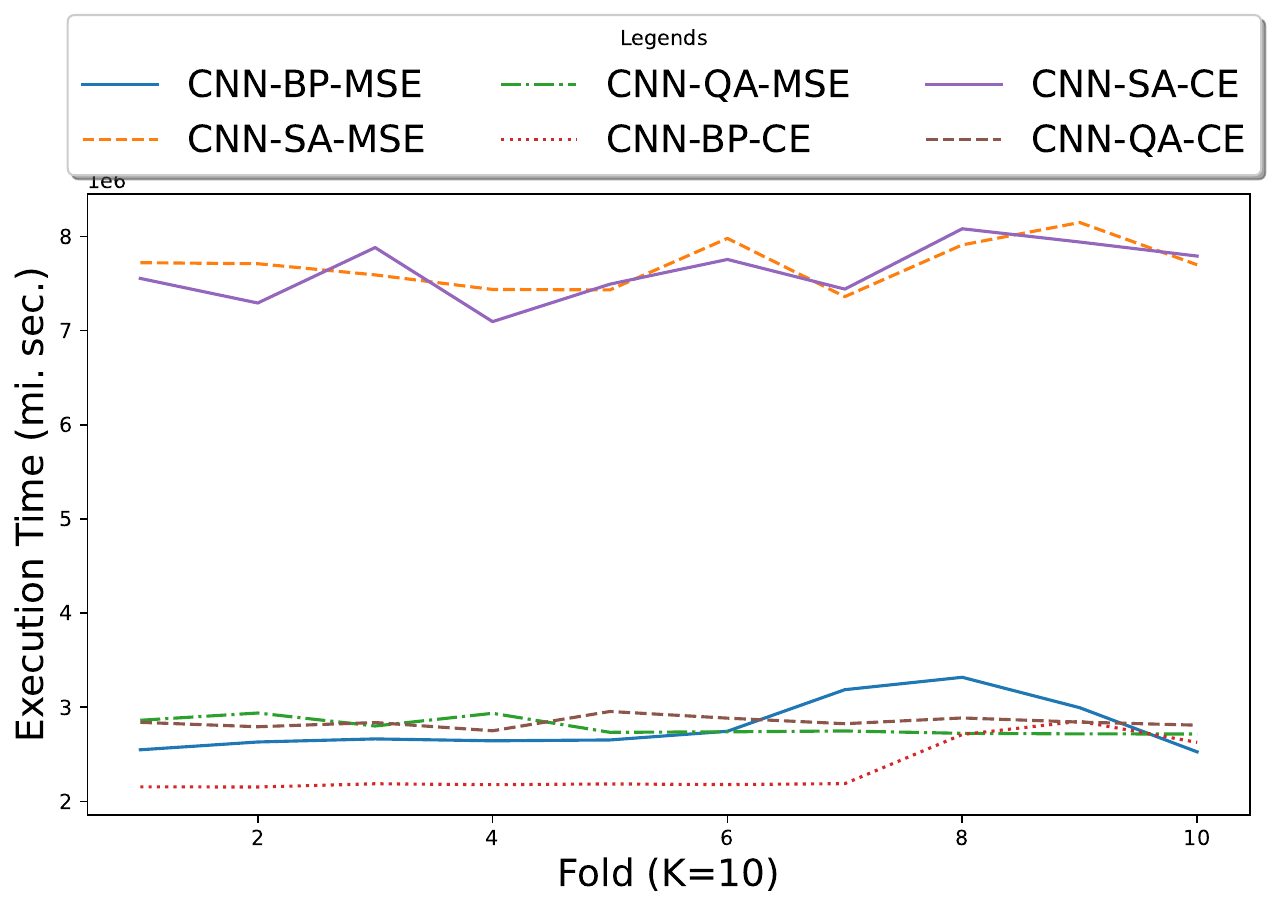}      
        \includegraphics[width=0.45\textwidth]{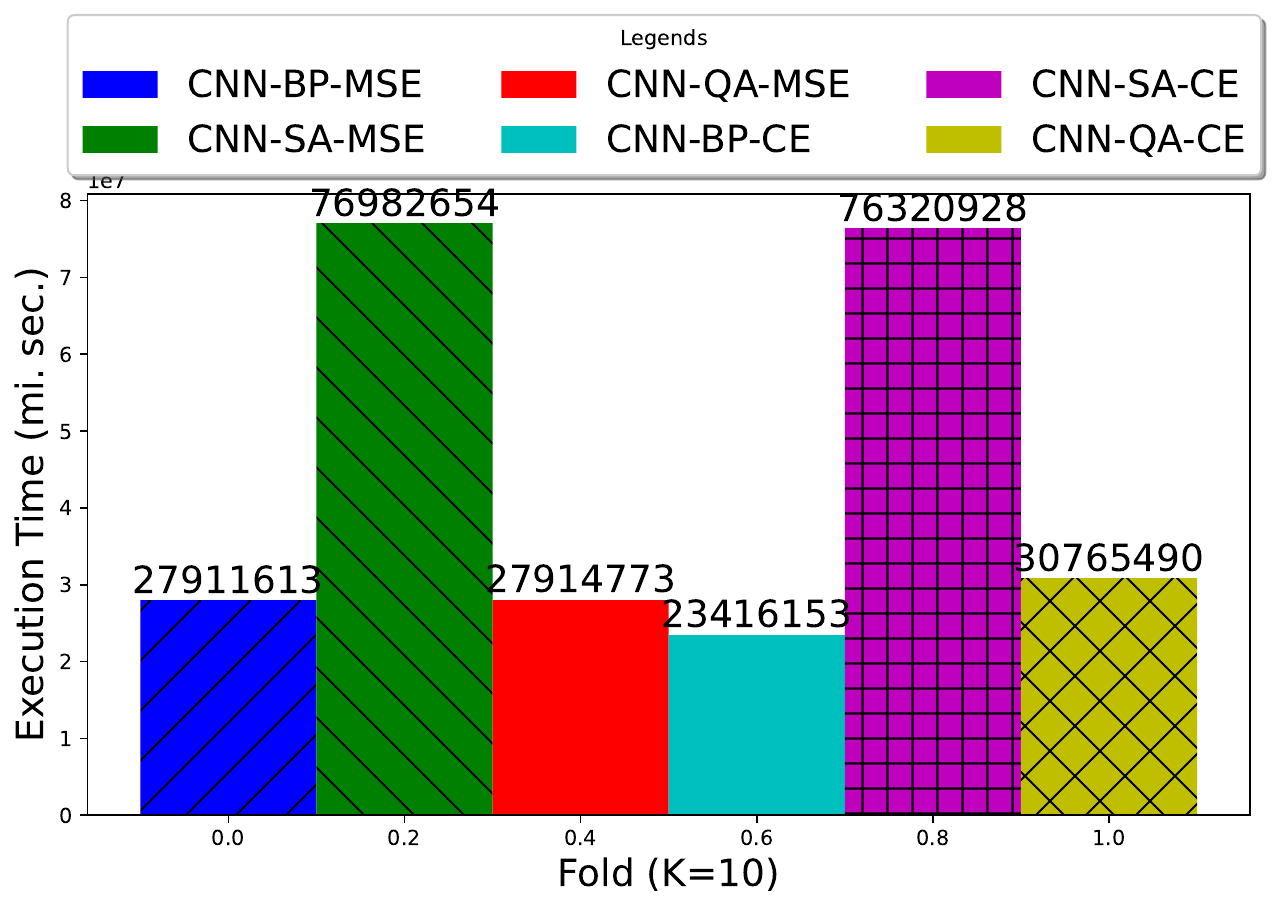}
        \caption{Execution Time (milliseconds)}
    \end{subfigure}
    
    \caption{Details of K=10 Fold Cross Validation Test}
    \label{fig:K10_FoldCrossValidationTestDetails}
\end{figure}

\begin{figure}
    \centering
    \begin{subfigure}[b]{0.5\textwidth}
        \includegraphics[width=0.45\textwidth]{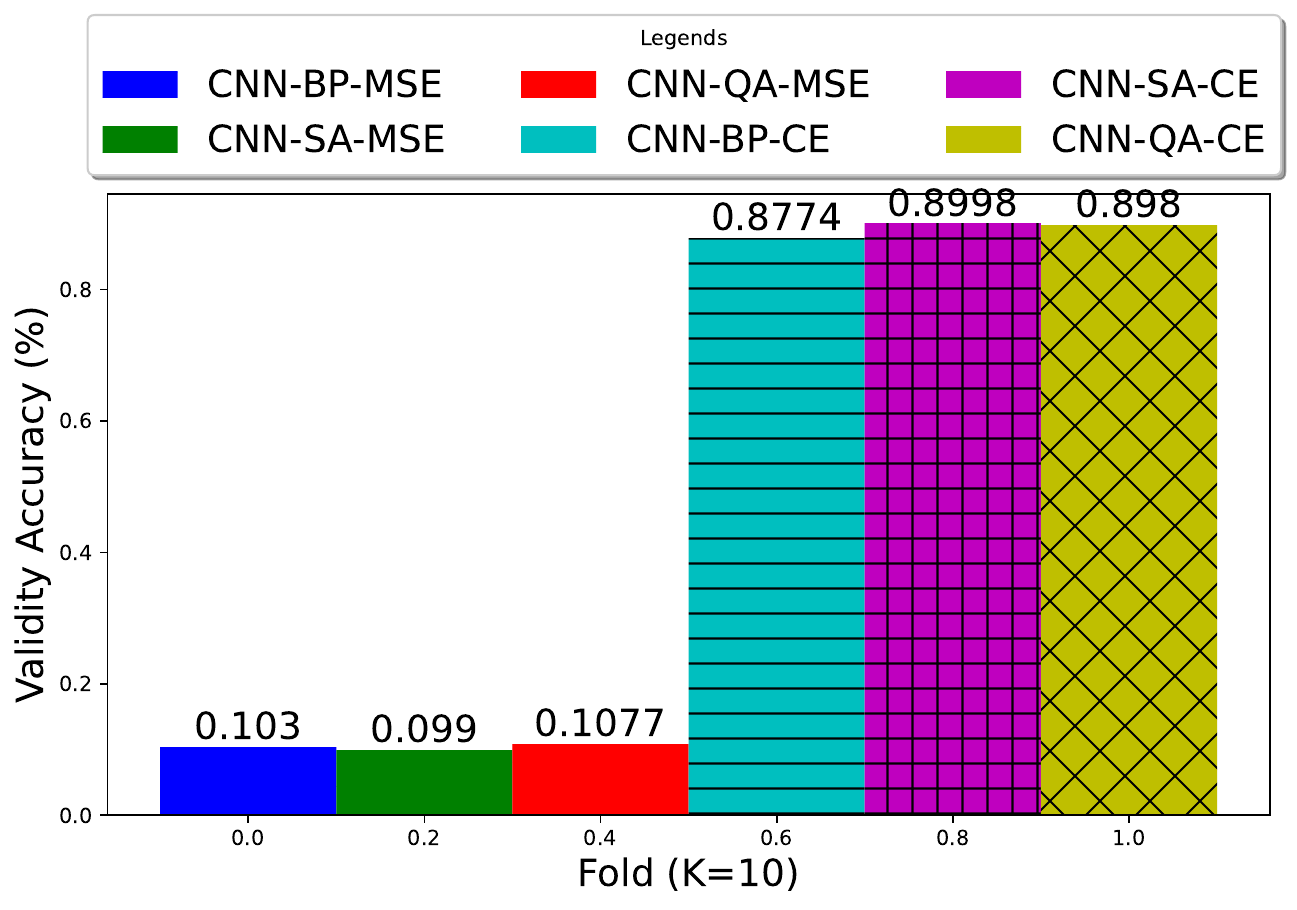} 
        \includegraphics[width=0.45\textwidth]{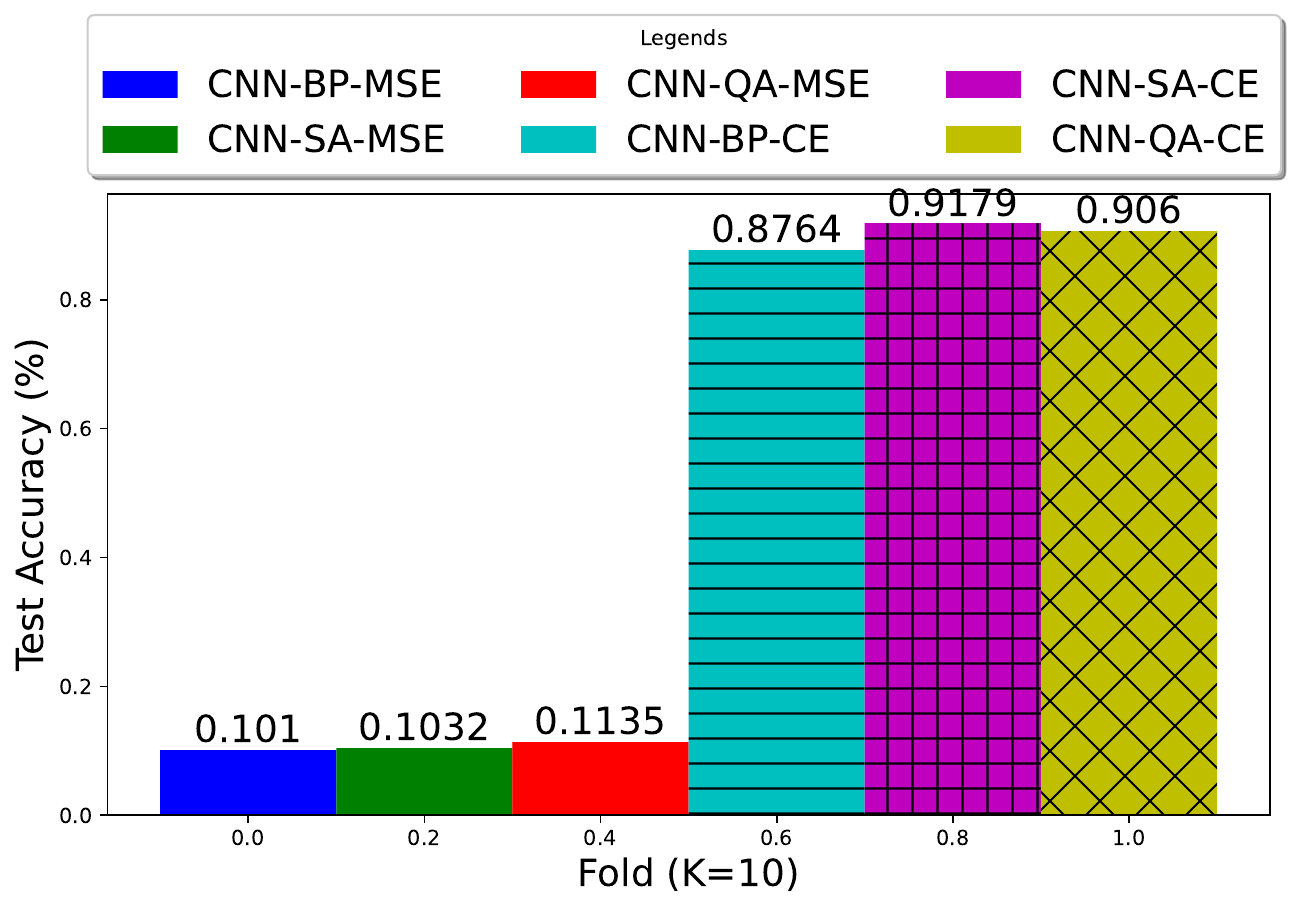}
        \caption{Validation and Test Accuracy (\%)}
    \end{subfigure}
    
    \begin{subfigure}[b]{0.5\textwidth}
        \centering
        \includegraphics[width=0.45\textwidth]{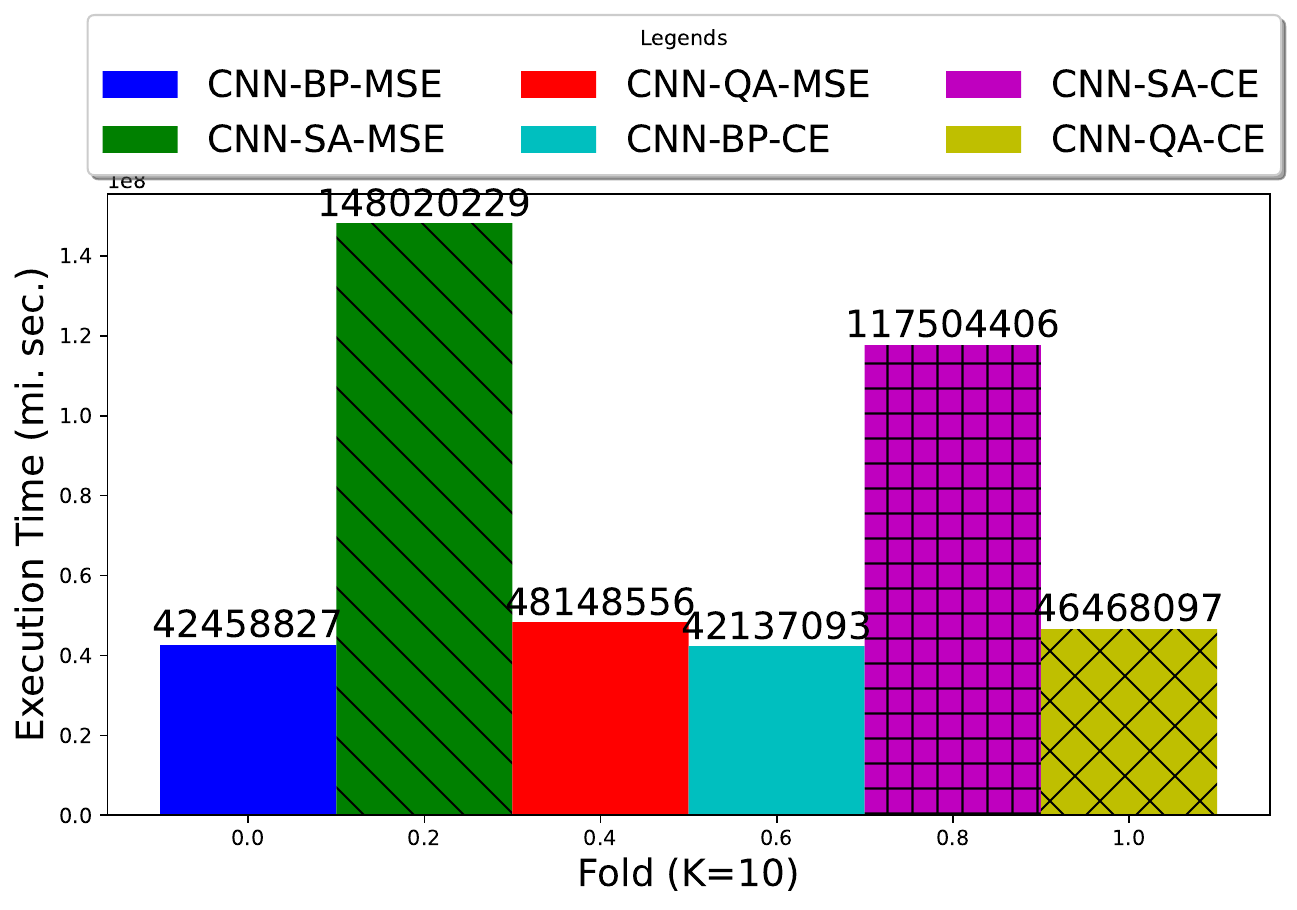}
        \caption{Execution Time (milliseconds)}
    \end{subfigure}
    
    \caption{Details of the 10 Epoch Test}
    \label{fig:10EpochTestDetails}
\end{figure}

\begin{figure}
    \centering
    \begin{subfigure}[b]{0.5\textwidth}
        \includegraphics[width=0.45\textwidth]{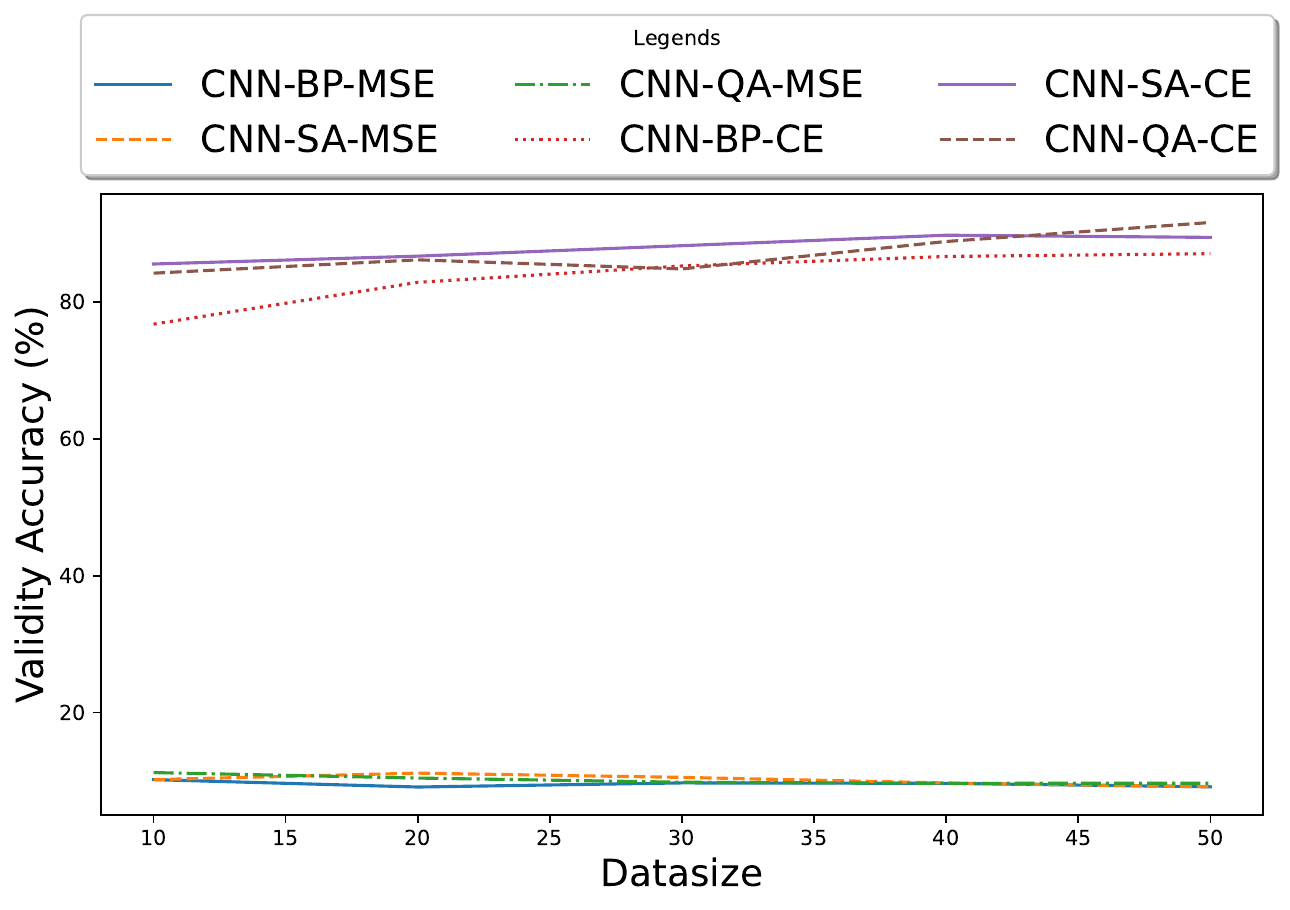}    
        \includegraphics[width=0.45\textwidth]{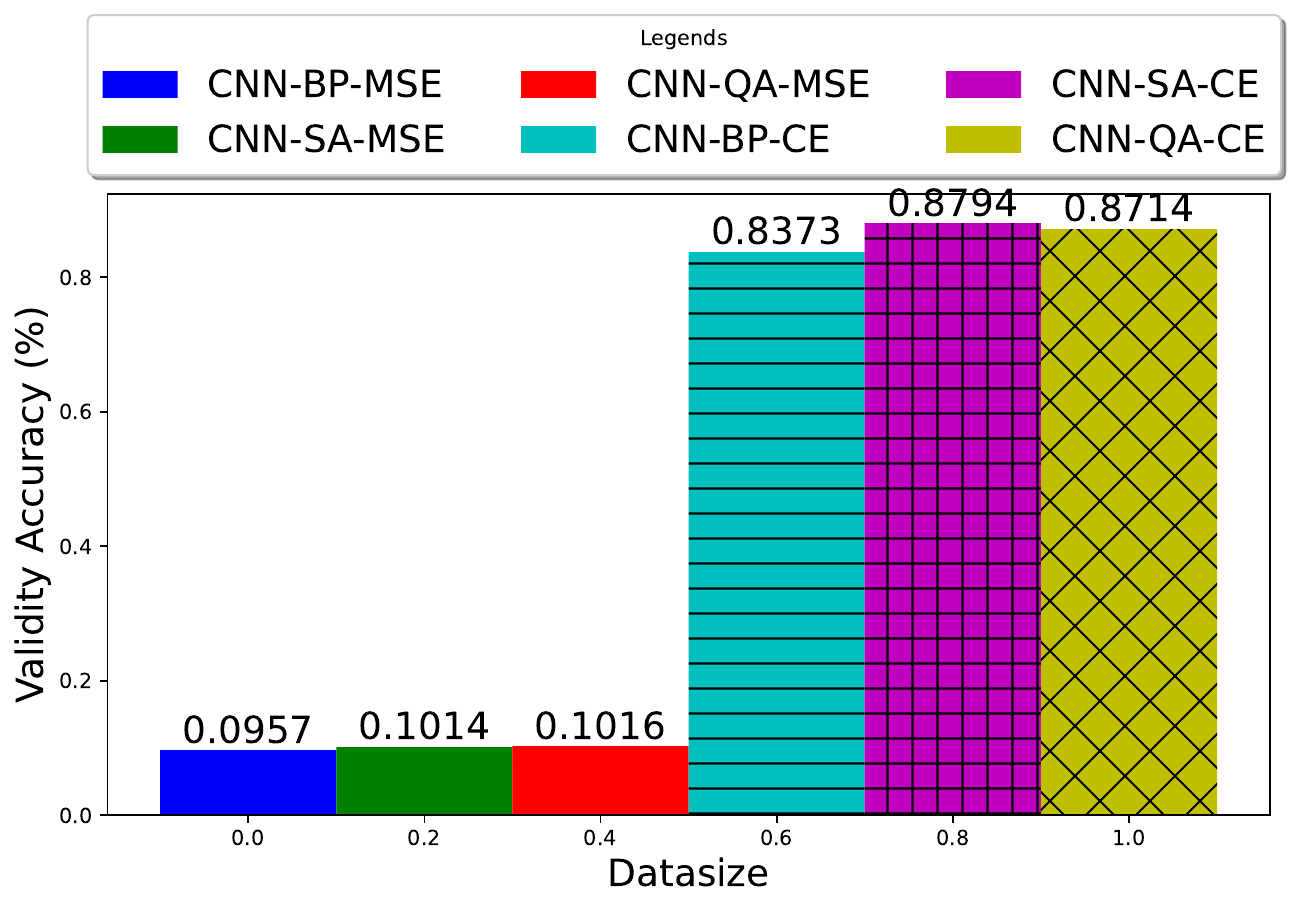} 
        \caption{Validation Accuracy (\%)}
    \end{subfigure}

    \begin{subfigure}[b]{0.5\textwidth}
        \includegraphics[width=0.45\textwidth]{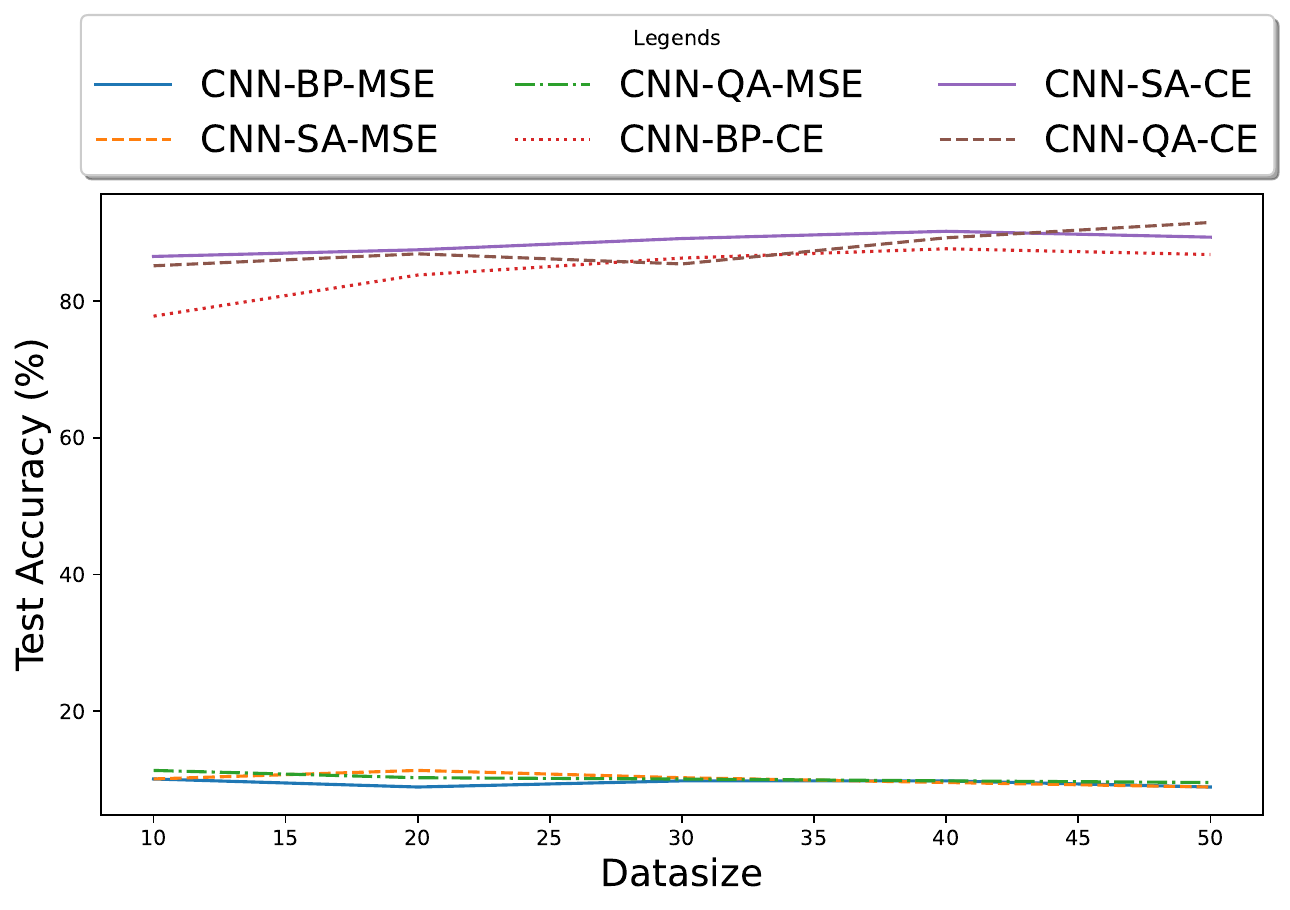}
        \includegraphics[width=0.45\textwidth]{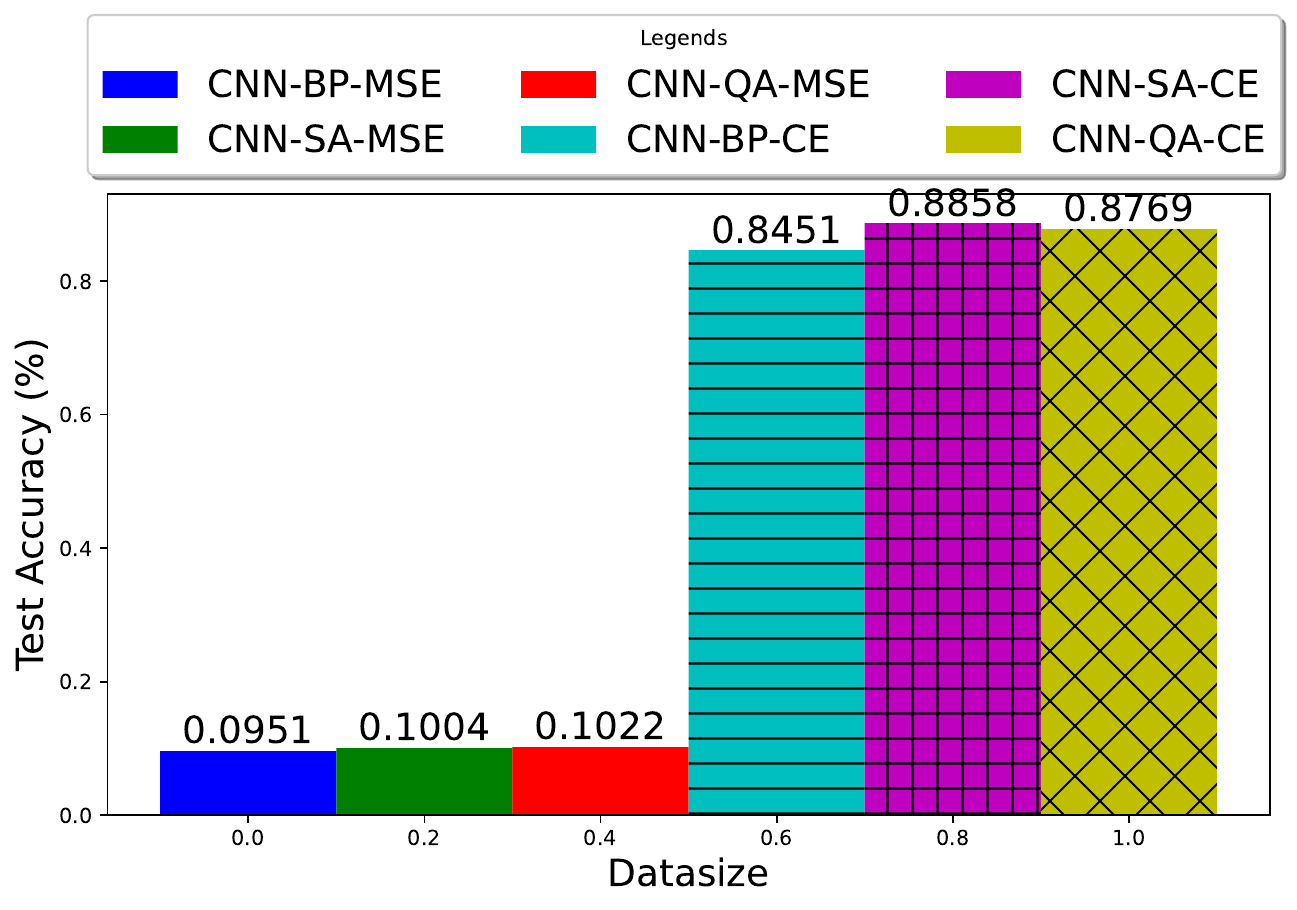}
        \caption{Test Accuracy (\%)}
    \end{subfigure}
    
    \begin{subfigure}[b]{0.5\textwidth}
        \includegraphics[width=0.45\textwidth]{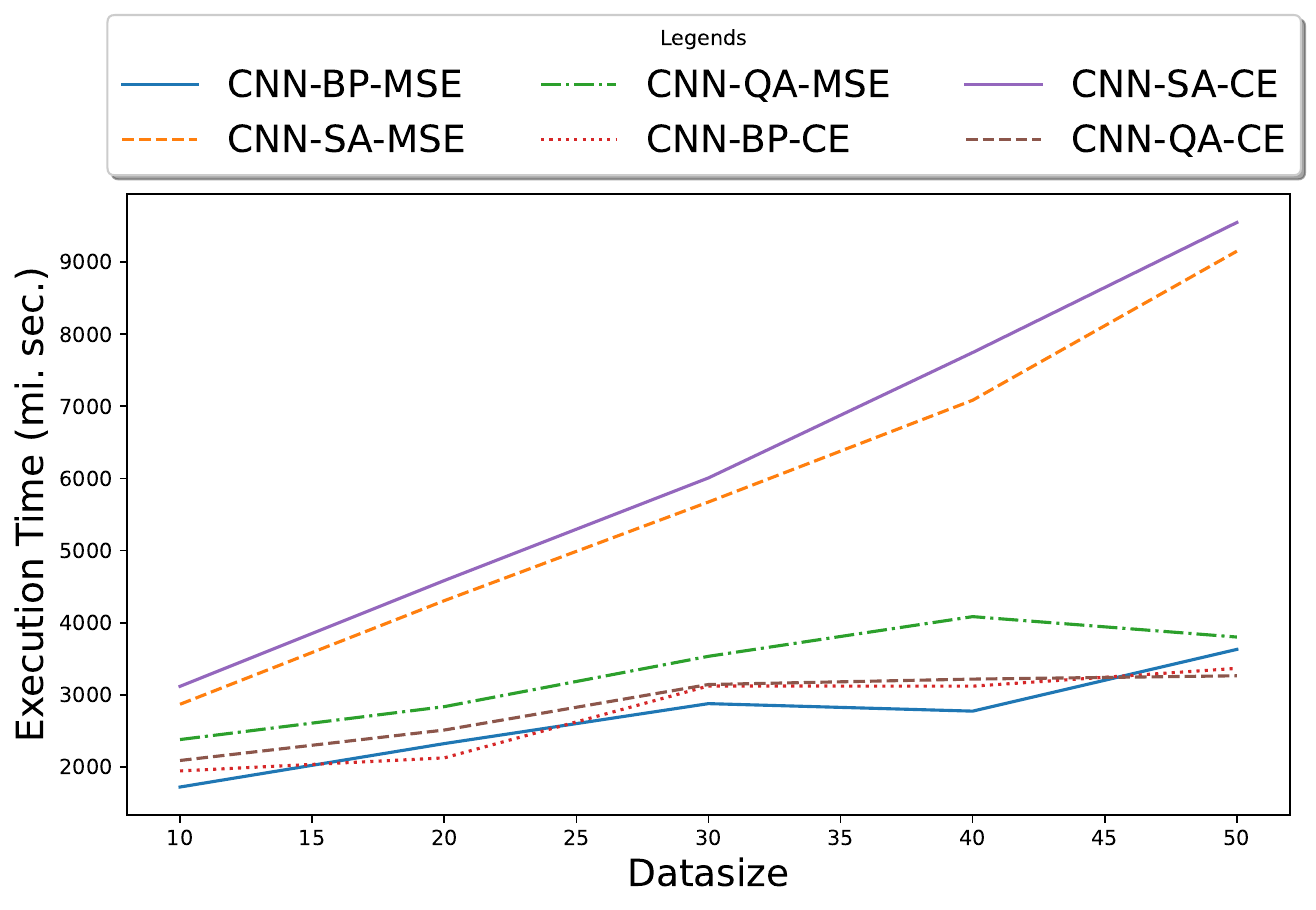}  
        \includegraphics[width=0.45\textwidth]{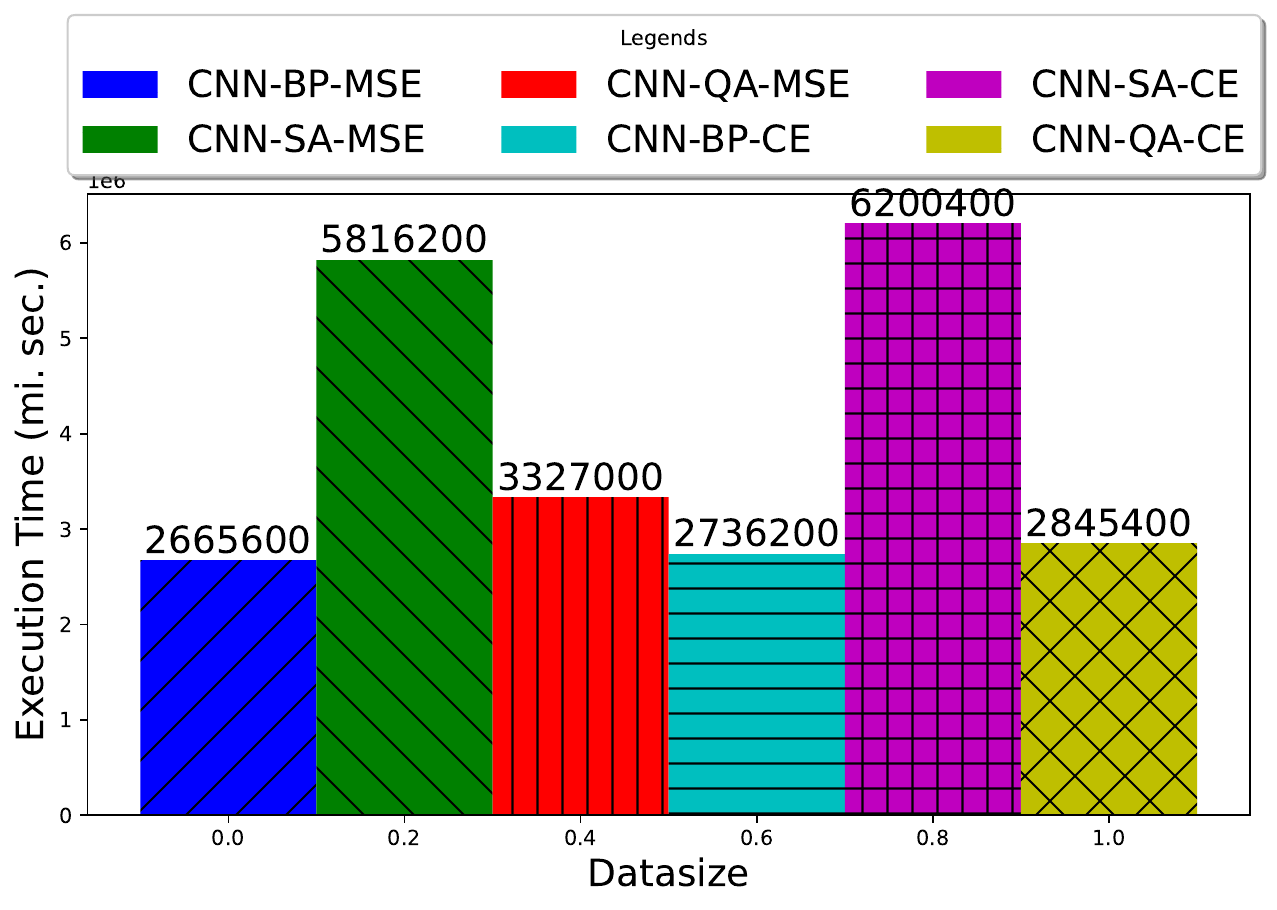}   
        \caption{Execution Time (milliseconds)}
    \end{subfigure}
    
    \caption{Cumulative Data Increment Test Details}
    \label{fig:CumulativeDataIncrementTest}
\end{figure}

The subsequent contents in \Cref{fig:K10_FoldCrossValidationTestDetails}, \Cref{fig:10EpochTestDetails}, and \Cref{fig:CumulativeDataIncrementTest} outlines the observations and procedures concerning the training and testing of the experimental models.
\par

\subsection{Results and Discussion}

This section presents the consolidated results from the series of tests executed, accompanied by a discussion detailing each validation and verification process undertaken.

\subsubsection{10-Fold Cross-Validation Test}

The 10-Fold Cross-Validation test results displayed in \Cref{fig:K10_FoldCrossValidationTestDetails} represent how CNN-BP, CNN-SA, and CNN-QA models perform given a learning rate (LR) of 0.1 using both CE and MSE objective functions. The analyses yielded the following observations:

\begin{itemize}
    \item The CNN-QA model exhibited superior performance, achieving higher accuracy by 40\% for MSE and by 6\% for CE objective functions compared to CNN-BP and CNN-SA at LR 0.1. 
    \item CNN-QA exhibited the shortest execution times, comparable to those of CNN-SA, around 2.76 times compared to CNN-QA for both CE and MSE.
    \item Under the configured hyper-parameters, CE consistently outperformed MSE in terms of accuracy across all models. Yet, CNN-QA produced higher accuracy for MSE, possibly leveraging its inherent advantages in search space evaluations.
    \item Overall, across both MSE and CE objectives, CNN-QA presented improved accuracy around 10-15\% and faster execution times.
\end{itemize}

\subsubsection{10 Epochs Test}

The 10-Epoch Test (\Cref{fig:10EpochTestDetails}) conducted with a LR of 0.1 for each of the CNN-BP, CNN-SA, and CNN-QA models, revealed:

\begin{itemize}
    \item CNN-QA demonstrated competitive performance against CNN-SA and CNN-BP in both validation and test accuracy for the MSE and CE objectives. 
    \item CNN-SA displayed 2.79 times longer execution times compared to both CNN-BP and CNN-QA, which had similar times.
    \item Overall, CNN-QA yielded competitive accuracy with shortened execution times across the 10 epoch test.
\end{itemize}

\subsubsection{Cumulative Data Increment Test}

This test, described in detail in \Cref{fig:CumulativeDataIncrementTest}, evaluated subset partitions of the dataset using a single epoch and LR of 0.1 for both objective functions. The findings include:

\begin{itemize}
    \item For both objectives, model performances in terms of accuracy were similar among the three models; however, they displayed linear growth in execution times, with the exception of the CNN-QA model.
    \item For the MSE objective, all models showed similar trends in accuracy with minor deviations at smaller data sizes. 
    \item CNN-BP and CNN-QA maintained steady execution times, whereas CNN-SA times increased substantially with data size. 
\end{itemize}

\section{Conclusion and Recommendation}
\label{sec:ConclusionAndRecommendations}
\subsection{Conclusion}
The objective of this research was to find a technique for DL, like CNN-BP that improves execution time and, akin to CNN-SA, enhances accuracy. The test results indicate that the QA-inspired CNN-QA model could successfully meet these criteria.

\begin{itemize}
\item \textit{Hybrid CNN Model (CNN-QA) Approach:}
As observed in the results and discussion section (\ref{sec:AnalysisResultAndDiscussion}), the hybrid CNN-QA model showed promising results. The QA technique resulted in significantly optimized training time in comparison to the SA method and reduced time complexity, enabling less iteration-focused optimization. However, this efficiency is contingent on the specific objective functions used.
\item \textit{Key Observations:}
The hybrid-QA strategy expectedly optimized the use of qubits compared to other QUG and QAOA strategies, while not compromising performance or accuracy.
\item \textit{Limitations:}
The study faced constraints related to the nature and number of tests. For example, the MSE function, aligning with the nature of combinatorial optimization, exhibited superior performance over the logarithmic CE objective function, which is not inherently compatible with combinatorial optimization. The study was further limited by encompassing just two objective functions and only one standard MNIST dataset using a simulator, with minimal testing on DWAVE's real machine (Leap).
\end{itemize}

\subsection{Recommendation: Implications and Future Directions}
The hybrid CNN-QA model represents a promising domain for future research. In the HPC systems context, such a model could dramatically reduce execution time for large-scale DL analyses while utilizing existing classical resources. Future research may explore expanding the techniques deployed in this study, such as QA, which could reveal more intricate relationships between QA and specific optimization problems. 

The learnings from working with emerging technologies like this contribute to advancing the optimization of complex problems and offer pragmatic and scalable solutions. Therefore, a continued pursuit of these technologies is imperative.

\section*{Acknowledgment}
The authors sincerely thank theGesellschaft für wissenschaftliche Datenverarbeitung mbH Göttingen (GWDG, Germany), and NCIT College (Pokhara University, Nepal) for their valuable contributions. We are especially grateful to D-Wave for access to the Leap Quantum Annealer, which supported key experiments. We also acknowledge the support from NHR\footnote{\url{http://www.nhr-verein.de/en/ourpartners}}, essential for presenting this work, and appreciate the constructive feedback from our peers.

This research was funded by the EU KISSKI Project under grant number 01|S22093A (Förderkennzeichen).
The implementation source code is available at \url{https://github.com/AasishKumarSharma/CNNApplyingUBQP}.

\bibliographystyle{IEEEtran} 
\bibliography{references}

\appendix

\end{document}